\documentclass[10pt,journal,compsoc]{IEEEtran}
\usepackage{grffile}
\usepackage{amsmath,amssymb} 
\usepackage{algorithm}
\usepackage{algorithmic}
\usepackage[margin=0pt,font=small,labelfont=small,justification=justified,labelsep=period]{caption}
\usepackage{bm}
\usepackage{multirow}
\usepackage{booktabs}
\usepackage{array}
\usepackage{mathtools}
\usepackage{color}
\usepackage{xspace}
\usepackage{color}
\usepackage{multirow}
\usepackage{graphics}
\usepackage{adjustbox}
\usepackage{subfigure}
\usepackage{amsmath}
\usepackage{paralist} 
\usepackage{enumitem} 
\usepackage{booktabs} 
\usepackage{array}

\graphicspath{{figures/}}

\usepackage[pagebackref=true,breaklinks=true,letterpaper=true,colorlinks,citecolor=blue,linkcolor=blue,bookmarks=false]{hyperref}

\newcommand{\Paragraph}[1]{{\flushleft\textbf{#1}}} 

\long\def\ignorethis#1{}

\def\first#1{\red{\textbf{#1}}}
\def\second#1{\blue{\underline{#1}}}
\usepackage{tabularx}

\def\modify#1{{{#1}}}
	{\unskip''\end{quotation}}

\definecolor{gray}{rgb}{0.5,0.5,0.5}
\definecolor{MyBlue}{rgb}{0,0,1.0}
\definecolor{MyYellow}{rgb}{0.9,0.9,0}
\definecolor{MyRed}{rgb}{0.8,0.2,0}
\definecolor{MyGreen}{rgb}{0,0.5,0.0}
\definecolor{MyGray}{rgb}{0.4,0.4,0.4}

\def\red#1{\textcolor{MyRed}{#1}}
\def\blue#1{\textcolor{MyBlue}{#1}}

\def\first#1{\red{\textbf{#1}}}
\def\second#1{\blue{\underline{#1}}}

\newlength\paramargin
\newlength\figmargin
\newlength\secmargin

\newcolumntype{L}[1]{>{\raggedright\let\newline\\\arraybackslash\hspace{0pt}}m{#1}}
\newcolumntype{C}[1]{>{\centering\let\newline\\\arraybackslash\hspace{0pt}}m{#1}}
\newcolumntype{R}[1]{>{\raggedleft\let\newline\\\arraybackslash\hspace{0pt}}m{#1}}

\def\eg{e.g.,~}
\def\etc{etc}

\def\etal{et~al.\xspace}

\newcommand{\secref}[1]{Section~\ref{sec:#1}}
\newcommand{\figref}[1]{Fig.~\ref{fig:#1}}
\newcommand{\tabref}[1]{Table~\ref{tab:#1}}

\ifCLASSOPTIONcompsoc
  \usepackage[nocompress]{cite}
\else
  \usepackage{cite}
\fi  

\begin{document}
\title{MEMC-Net: Motion Estimation and Motion Compensation Driven Neural Network for \\ Video Interpolation and Enhancement}
\def\Ours{MEMC-Net}
\def\lastNet{post-processing\ }
\def\LastNet{Post-processing}
\def\lastNetShort{post-proc.}
\author{Wenbo~Bao,~
        Wei-Sheng~Lai,~
        Xiaoyun~Zhang,~
        Zhiyong~Gao,~
        and~Ming-Hsuan~Yang
\IEEEcompsocitemizethanks{\IEEEcompsocthanksitem Wenbo Bao, Xiaoyun Zhang and Zhiyong Gao are with the Department of Electrical Engineering, Shanghai Jiao Tong University, Shanghai, 200240. Email: \{baowenbo$\vert$xiaoyun.zhang$\vert$zhiyong.gao\}@sjtu.edu.cn
\IEEEcompsocthanksitem Wei-Sheng Lai and Ming-Hsuan Yang are with the Department of Electrical Engineering and Computer Science, University of California, Merced, CA, 95340. Email:\{wlai24$\vert$mhyang\}@ucmerced.edu}
}

 
\markboth{IEEE Transactions on Pattern Analysis and Machine Intelligence}{}

\IEEEtitleabstractindextext{%
\begin{abstract}
Motion estimation (ME) and motion compensation (MC) have been widely used for classical video frame interpolation systems over the past decades.
Recently, a number of data-driven frame interpolation methods based on convolutional neural networks have been proposed.
However, existing learning based methods typically estimate either flow or compensation kernels, thereby limiting performance on both computational efficiency and interpolation accuracy.
In this work, we propose a motion estimation and compensation driven neural network for video frame interpolation.
A novel adaptive warping layer is developed to integrate both optical flow and interpolation kernels to synthesize target frame pixels.
This layer is fully differentiable such that both the flow and kernel estimation networks can be optimized jointly.
The proposed model benefits from the advantages of motion estimation and compensation methods without using hand-crafted features.
%
Compared to existing methods, our approach is computationally efficient and able to generate more visually appealing results.
Furthermore, the proposed MEMC-Net architecture can be seamlessly adapted to several video enhancement tasks, e.g., super-resolution, denoising, and deblocking.
Extensive quantitative and qualitative evaluations demonstrate that the proposed method performs favorably against the state-of-the-art video frame interpolation and enhancement algorithms on a wide range of datasets.

\end{abstract}

}

\maketitle

\IEEEdisplaynontitleabstractindextext

%
\IEEEpeerreviewmaketitle

\IEEEraisesectionheading{\section{Introduction}\label{sec:introduction}}

%
\IEEEPARstart{V}{ideo} frame interpolation aims to synthesize non-existent frames between original input frames, which has been applied to numerous applications such as video frame rate conversion~\cite{castagno1996method}, novel view synthesis~\cite{flynn2016deepstereo}, 
and frame recovery in video streaming~\cite{wu2016modeling}, to name a few.
%
Conventional approaches~\cite{de1993true,bao2018high} are generally based on motion estimation and motion compensation (MEMC), and have been widely used in various display devices~\cite{wu2015enabling}.
A few deep learning based frame interpolation approaches~\cite{long2016learning,mathieu2015deep} have been developed to address this classical topic.
In this paper, we analyze the MEMC-based and learning-based approaches of video frame interpolation and exploit the merits of both paradigms to propose a high-quality frame interpolation processing algorithm. 

\begin{figure*}[t]
	\footnotesize
	\centering
	\renewcommand{\tabcolsep}{2pt} 
	\renewcommand{\arraystretch}{1} 
	\begin{center}
		\begin{tabular}{ccccc}
\includegraphics[width=0.19\linewidth]{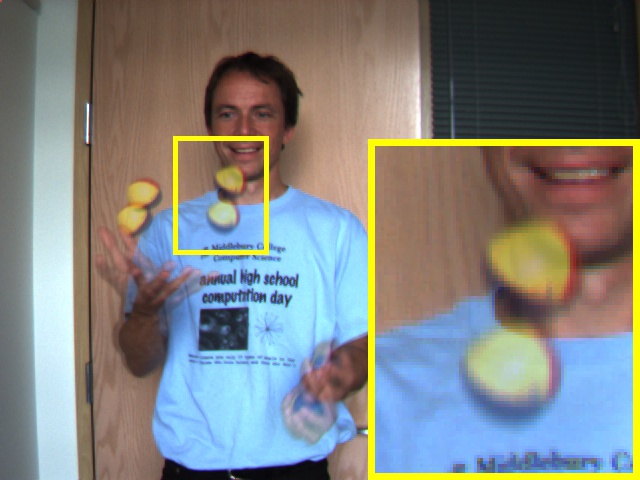}&		
\includegraphics[width=0.19\linewidth]{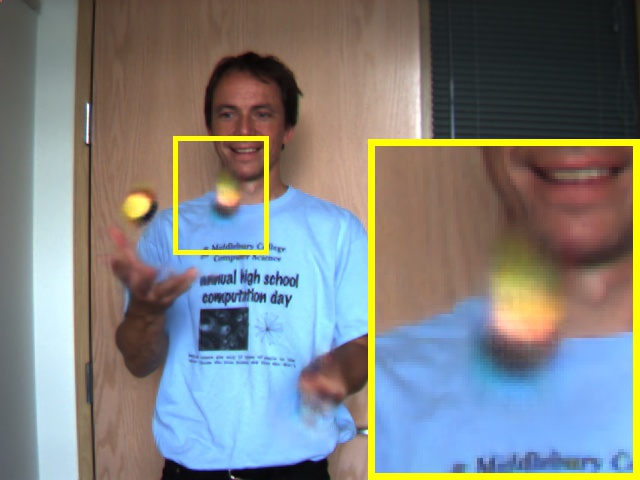}&
\includegraphics[width=0.19\linewidth]{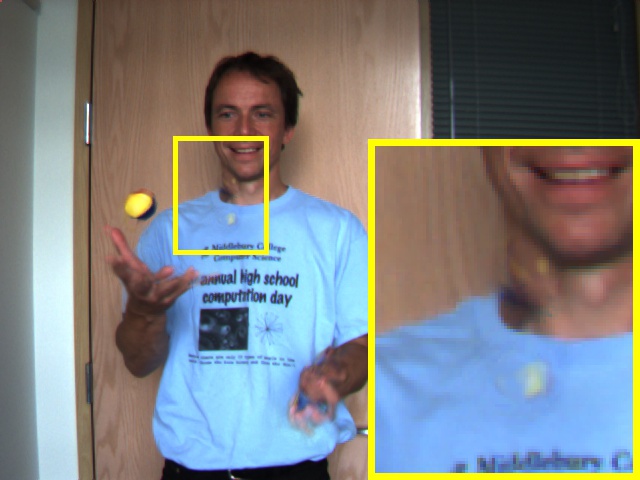}&
\includegraphics[width=0.19\linewidth]{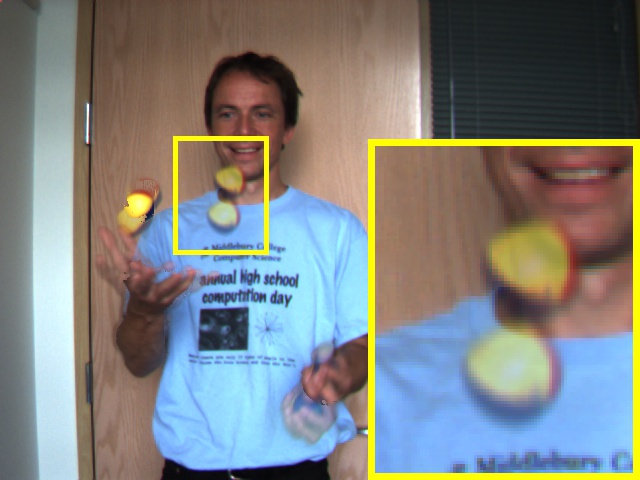}&
\includegraphics[width=0.19\linewidth]{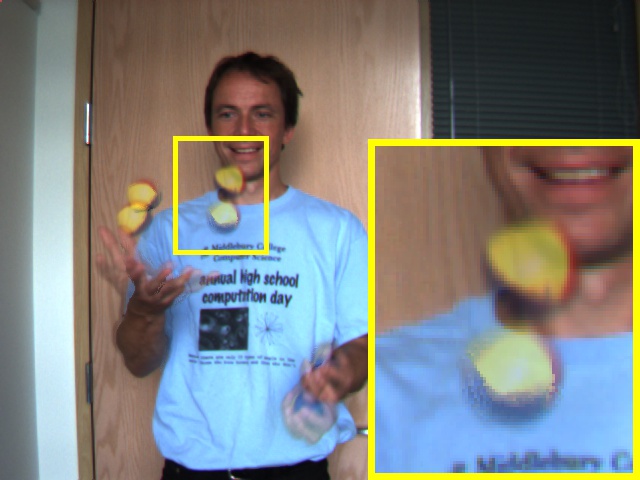}
 \\
(a) Overlay &
(b) MIND~\cite{long2016learning}&
(c) ToFlow~\cite{xue2017video}&
(d) EpicFlow~\cite{revaud2015epicflow}&
(e) SPyNet~\cite{ranjan2017optical} 
\\

\includegraphics[width=0.19\linewidth]{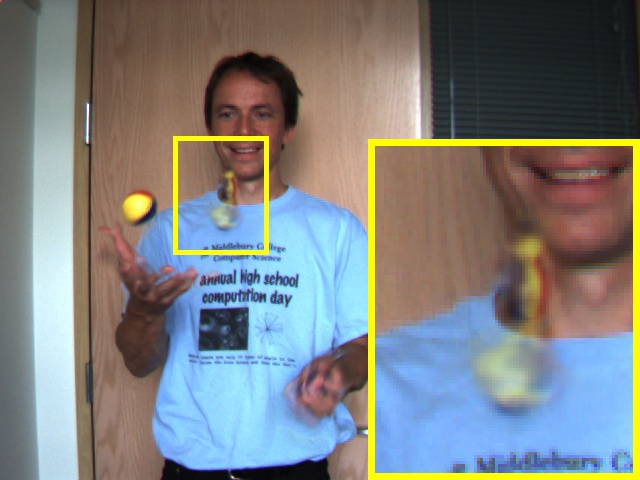}&
\includegraphics[width=0.19\linewidth]{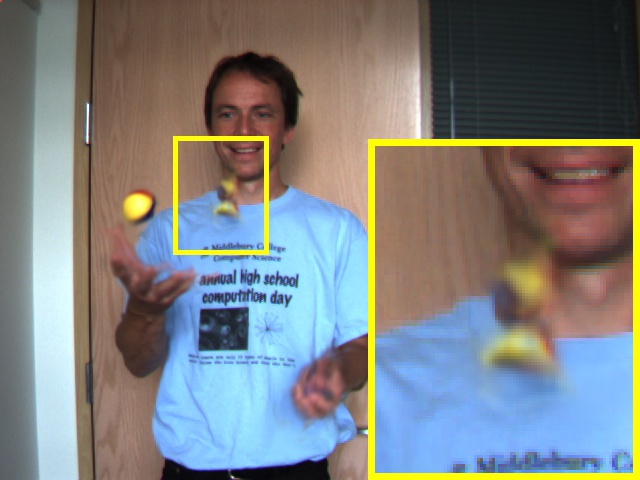}&
\includegraphics[width=0.19\linewidth]{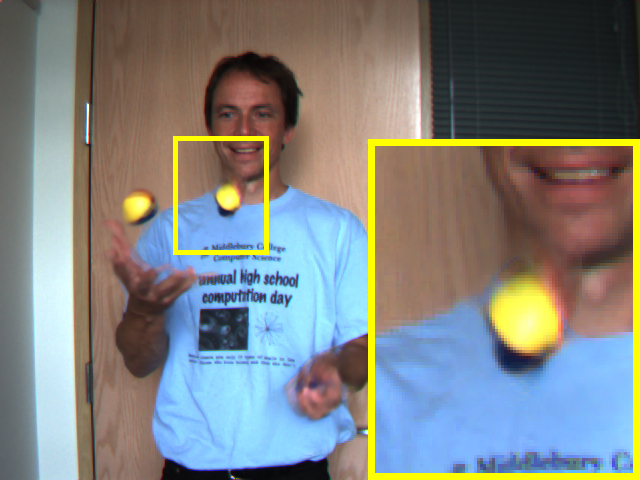}&
\includegraphics[width=0.19\linewidth]{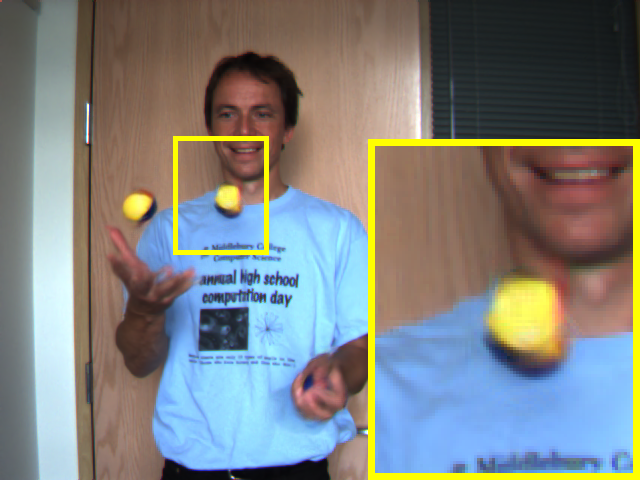}&
\includegraphics[width=0.19\linewidth]{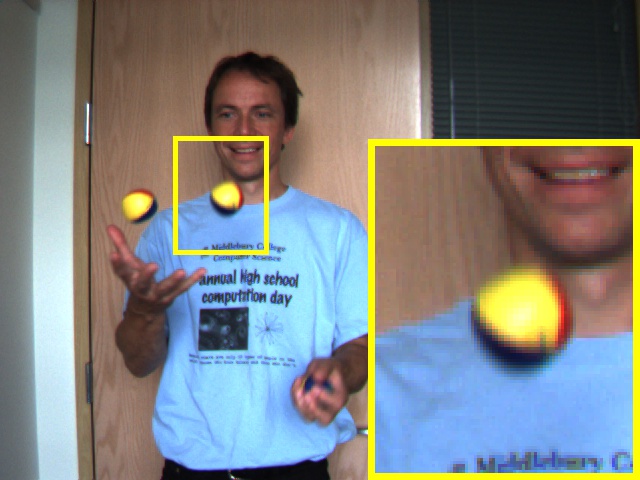} 
\\
			 
(f) SepConv-$L_f$~\cite{niklaus2017videoSepConv}&
(g) SepConv-$L_1$~\cite{niklaus2017videoSepConv}&
(h) \Ours~ &
(i) \Ours*~ &
(j) Ground Truth\\	
		\end{tabular}
	\end{center}
	\vspace{-0.5cm}
	\caption{
		\textbf{Visual comparisons with existing frame interpolation approaches.}
		The proposed method \Ours{} synthesizes the intermediate frame with clear edges and shape. 
		With the context information and residual blocks used, the improved model \Ours* obtains better outcome with fine details around motion boundaries.
	}
\label{fig:preface-compare} 
\end{figure*}

Conventional MEMC-based approaches entail both motion estimation~\cite{konrad1992bayesian} and motion compensation~\cite{orchard1994overlapped} for video interpolation. 
Motion estimation is used to determine the block-wise or pixel-wise motion vectors between two frames.
The block-based methods~\cite{de1993true} assume that the pixels within a block share the same motion and use search strategies~\cite{zhu2000new,gao2000multilevel} and selection criteria~\cite{de1993true,wang2010frame} to obtain the optimal motion vector.
On the other hand, the methods based on pixel-based motion, i.e., optical flow, estimate a motion/flow vector for each pixel of the frames and thus entail heavy computational loads.
The recent years have witnessed significant advances in optical flow estimation  via variational optimization~\cite{brox2004high}, nearest neighbor field search~\cite{chen2013large}, cost volume filtering~\cite{xu2017accurate}, and deep convolutional neural networks (CNNs)~\cite{dosovitskiy2015flownet,ranjan2017optical}.
However, estimating optical flow remains a challenging problem due to fast-moving and thin objects, occlusion and dis-occlusion, brightness change and motion blur.
To account for inaccurate flow and occluded pixels, motion compensated interpolation methods usually use sophisticated filters to reduce visual artifacts of the generated frames~\cite{orchard1994overlapped,choi2007motion}.
%
In addition, these schemes do not perform well where objects in the intermediate frame are invisible in both forward and backward reference frames (e.g., some pixels cannot be compensated), and 
require further post-processing procedures to fill in missing or remove unreliable pixels~\cite{biswas2010handling,lee2014frame,wang2010frame,kim2014new}. 
%

Numerous learning-based frame interpolation methods based on deep CNNs have been recently proposed~\cite{mathieu2015deep,long2016learning}.
%
The training datasets for learning-based methods typically contain image triplets from raw video sequences, with the first and third frame feeding into the network as inputs and the intermediate second frame acting as ground truth~\cite{mathieu2015deep,long2016learning,xue2017video} for output.
By imposing loss functions such as $L_p$-norm on the difference between the network output and ground truth frame pixels, the model parameters can be iteratively updated via a gradient descent scheme.
%


The conventional MEMC-based methods are computationally efficient due to the block-wise setting~\cite{zhai2005low,kim2014new}.
However, these block-based methods do not achieve the state-of-the-art results as hand-crafted features are typically used in the ME and MC stages.
%
%
In contrast, the learning-based methods are developed based on the massive amount of raw video data.
However, the state-of-the-art learning-based approaches~\cite{xue2017video,niklaus2018context} focus on motion estimation, which often leads to blurry results due to bilinear interpolation process. 
While other approaches~\cite{niklaus2017videoAdaConv,niklaus2017videoSepConv} are developed to consider the effect of interpolation kernels, such schemes are sensitive to large motion.
%
%

%
%


%
In this paper, we propose to exploit motion estimation and motion compensation in a neural network for video frame interpolation.
%
%
Both the motion vectors and compensation filters are estimated through CNNs.
We further propose an adaptive warping layer based on optical flow and compensation filters for synthesizing new pixels.
This novel warping layer is fully differentiable such that the gradients can be back-propagated to both the ME and MC networks.
To account for the occlusions, we estimate occlusion masks to adaptively blend the warped frames.
Furthermore, the missing pixels in holes and unreliable pixels of the warped frames are processed by a \lastNet CNN.
Our entire model, \Ours, is motivated by the architecture of conventional methods but realized via the most recent learning-based approaches.
\figref{preface-compare} shows an interpolated frame of our methods (\Ours\ and \Ours*) and existing algorithms~\cite{long2016learning,niklaus2017videoSepConv,xue2017video,revaud2015epicflow,ranjan2017optical}, where the proposed methods predict the moving ball with clearer contours and sharper edges.

The contributions of this paper are summarized as follows:
\begin{enumerate}[label=(\roman*)]
	\item[(1)] We propose a motion estimation and compensation driven neural network for robust and high-quality video frame interpolation.
	\item[(2)] We integrate the optical flow warping with learned compensation filters into an \emph{adaptive warping layer}.
	The proposed adaptive warping layer is fully differentiable and applicable to several video processing tasks, \eg video super-resolution, video denoising, and video deblocking.
	\item[(3)] We demonstrate that the proposed method performs favorably against the state-of-the-art frame interpolation algorithms on several benchmark datasets, including the Middlebury~\cite{baker2011database}, UCF101~\cite{soomro2012ucf101}, and Vimeo90K~\cite{xue2017video} datasets.
	Our model requires less memory to predict the compensation filters and executes efficiently.
	\item[(4)] We extend our network to the other video enhancement tasks including super-resolution, denoising, and deblocking as the model is general and applicable to motion compensation based tasks.
	Our methods obtain more favorable results against the state-of-the-art algorithms on each of these tasks.
\end{enumerate}

	%
	%

	%

\section{Related Work}
\label{sec:RW}
In this section, we discuss the conventional MEMC-based and recent learning-based methods.

\subsection{Conventional MEMC-based Methods}
\begin{figure}[!t]
	\centering
		\renewcommand{\tabcolsep}{2pt} 
		\renewcommand{\arraystretch}{0.5} 
		\centering
				\begin{tabular}{cl} 
					\vspace{5pt}
	\multirow{1}{*}[1.25em]{(a)} &
	\includegraphics[height=0.75cm]{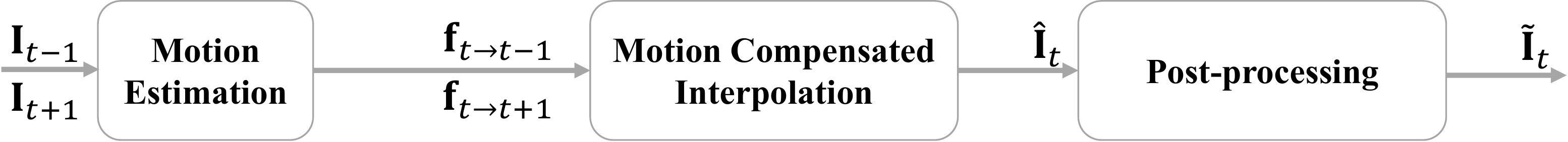}\\
	
	\vspace{5pt}
		
	\multirow{1}{*}[1.25em]{(b)} &
		\includegraphics[height=0.75cm]{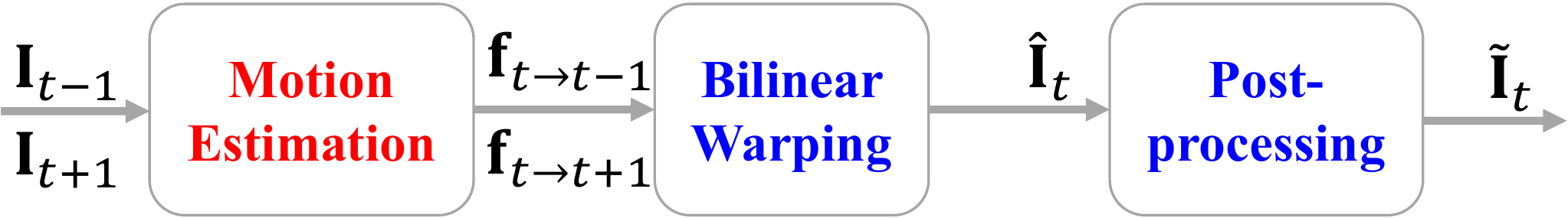}\\
    
   \vspace{5pt}
	
	\multirow{1}{*}[1.5em]{(c)} &
		\includegraphics[height=0.75cm]{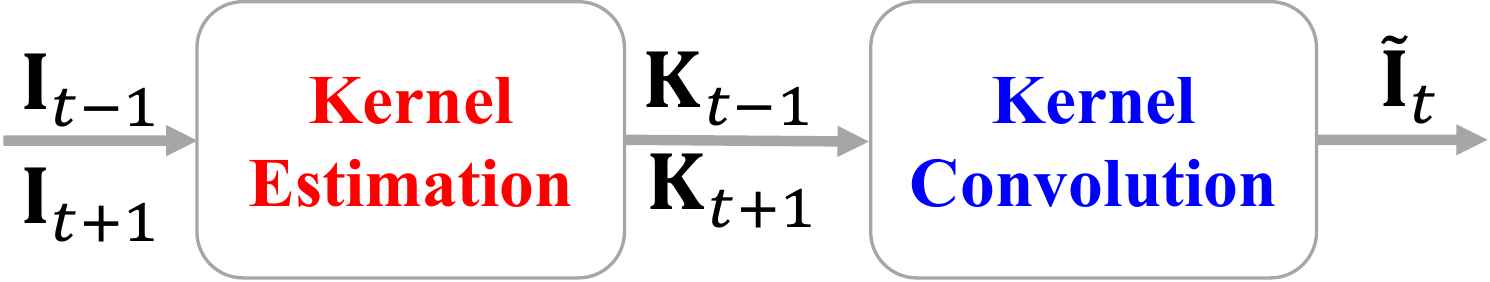}\\
	\end{tabular}
	
		\vspace{-0mm}
	\caption{
		\textbf{Frameworks of (a) the conventional MEMC-based approaches, (b) the flow-based and (c) the kernel-based models.}
	The \textbf{black}, \red{\textbf{red}}, and \blue{\textbf{blue}} text boxes correspond to the conventional modules, network modules, and network layers respectively. 
		}
	\label{fig:conventional_framework}
\end{figure}

\figref{conventional_framework}(a) shows the typical framework of conventional MEMC-based video frame interpolation methods.
First,  motion vectors between the forward and reference frames are estimated. 
Along the motion trajectories, pixels of the reference frames are used to interpolate the intermediate frame.
Conventional ME methods use block-based algorithms such as the 3D recursive search~\cite{de1993true}, which are hardware-friendly and computationally efficient.
The block-based methods typically divide the image frames into small pixel blocks and exploit certain search strategies such as spatial/temporal search~\cite{de1993true}, hierarchical
search~\cite{nam1995fast}, based on 
selection criteria such as the minimum sum of absolute block difference to compute their motion vectors.
%

%
For motion compensated interpolation,
overlapped blocks are usually utilized 
to cope with the erroneous motion vectors of pixel blocks~\cite{orchard1994overlapped}.
Recently, several methods~\cite{lee2014frame,kaviani2016frame} exploit
optical flow for the truthfulness of flow fields.
Compensation filters via image fusion~\cite{lee2014frame} or overlapped patch reconstruction~\cite{kaviani2016frame} are developed to deal with occlusion or blocky effects.

{
Aside from the motion estimation and motion compensation procedures, a \lastNet step is often required to minimize artifacts and improve visual qualities~\cite{wang2010frame,lee2014frame,biswas2010handling,kim2014new}.
Due to relative motions and occlusion between objects with different depth,  the estimated flow vectors may lead to incorrect interpolation results with hole regions. 
%
%
Kim~\etal~\cite{kim2014new} utilize a hole interpolation method to restore missing pixels.
On the other hand, Wang~\etal~\cite{wang2010frame} propose a trilateral filtering method to fill the holes and smooth the compensation errors in both the spatial and temporal domains.
}
The proposed algorithm differs from the conventional MEMC methods
in that we develop a data-driven end-to-end trainable model with deep features.
%
%

\subsection{Learning-based Methods}
Video frame interpolation based on deep learning algorithms can be categorized into the direct method, phase-based, flow-based, and kernel-based approaches.
Long~\etal~\cite{long2016learning} train a deep CNN to directly predict the interpolated frames.
The outputs are usually blurry and contain fewer details as this deep model is not able to capture the multi-modal distribution of natural images and videos.
The phase-based method~\cite{meyer2015phase} manipulates the pixel phase information within a multi-scale pyramid for frame interpolation.
However, this approach is less effective in handling large motion in complicated scenes.
In the following, we focus our discussion on recent flow-based and kernel-based methods.

\Paragraph{Flow-based methods.}
With the advances in optical flow estimation by deep CNNs~\cite{dosovitskiy2015flownet,ranjan2017optical,ilg2017flownet,lai2017semi}, several 
methods based on end-to-end deep models have been developed for frame interpolation.
These approaches either predict bi-directional flow~\cite{xue2017video} or 
use the bilinear warping operation to align input frames
based on linear motion models~\cite{liu2017video,van2017frame,jiang2017super}.
To synthesize an output image, a common technique is to estimate an occlusion mask 
to adaptively blend the warped frames.
As the bilinear warping blend neighbor pixels based on the sub-pixel shifts, the flow-based methods inevitably generate ghost or blurry artifacts when the input frames are not aligned well.
{The pipeline for the flow-based methods is illustrated in~\figref{conventional_framework}(b).}
Instead of using the fixed bilinear coefficients for interpolation, our approach learns spatially-varying interpolation kernels for each pixel.
The learned kernels have larger spatial support (\eg $4 \times 4$) than the bilinear interpolation and thus better account for occlusion and dis-occlusion.

\Paragraph{Kernel-based methods.}
Instead of relying on pixel-wise optical flow, frame interpolation can be formulated 
as convolution operations over local patches~\cite{zhang2007spatio,zhang2009spatio}.
Niklaus~\etal~\cite{niklaus2017videoAdaConv} propose the AdaConv model to estimate spatially-adaptive convolutional kernels for each output pixel.
{
We show the pipeline of kernel-based methods in~\figref{conventional_framework}(c).}
In these methods, a large kernel size is used to handle large motion, 
which requires a large amount of memory to process high-resolution images.
For an input frame of $H \times W$ pixels, the AdaConv model needs to 
estimate $H \times W \times R \times R$ coefficients for interpolation, where $R$ is the size of the local kernels.
To reduce memory requirements, the SepConv method~\cite{niklaus2017videoSepConv} assumes that the convolutional kernels are separable and uses a pair of 1D kernels (one vertical and one horizontal kernel) to approximate the 2D kernels.
This strategy significantly reduces the memory consumption from $O(R^2)$ to $O(2R)$ and further improves 
interpolation results. 
However, both the AdaConv and SepConv methods cannot handle motion larger than the \emph{pre-defined} kernel size.
While our approach also learns adaptive local kernels for interpolation, 
the proposed method is not limited by the assumption of \emph{fixed} motion range 
as optical flow warping is integrated.  
That is, our method uses smaller kernels, requires a low amount of memory, and performs robustly 
to frames with large motion.
We list the main difference with flow-based methods~\cite{liu2017video,xue2017video} and kernel-based approaches~\cite{niklaus2017videoAdaConv,niklaus2017videoSepConv} in~\tabref{compare_stoa}.

\begin{figure}[!t]
	\centering
	\renewcommand{\tabcolsep}{2pt} 
	\renewcommand{\arraystretch}{0.5} 
	\centering
	\begin{tabular}{cl} 
		\vspace{5pt}
		
		\vspace{5pt}
		
		\multirow{1}{*}[1.25em]{(a)} &
	\includegraphics[height=0.75cm]{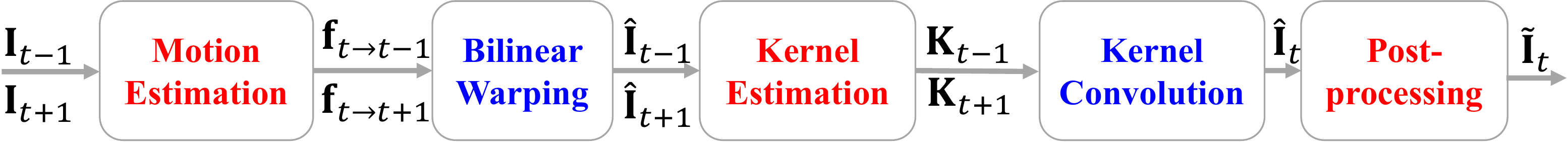}\\
		
		\vspace{5pt}
		
		\multirow{1}{*}[3.5em]{(b)} &
		\includegraphics[height=1.57cm]{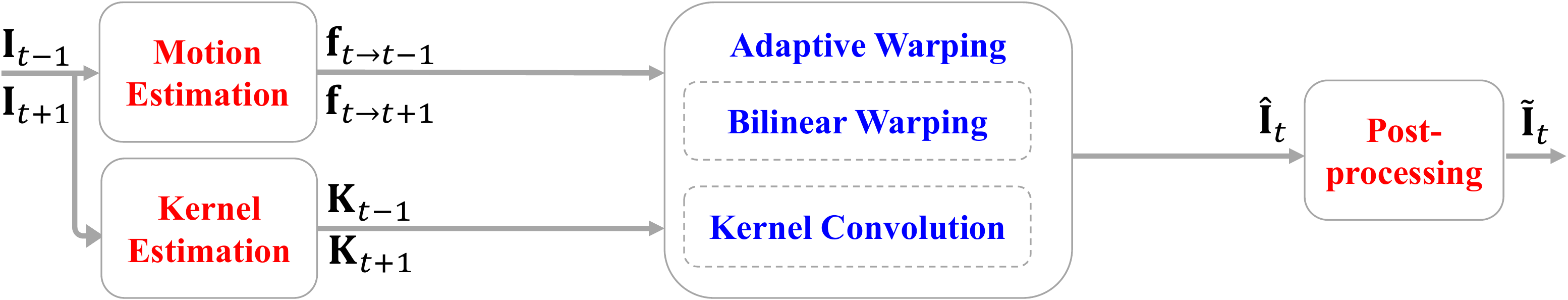}\\
	\end{tabular}
	
	\vspace{-0mm}
	\caption{
		\textbf{Frameworks of (a) the sequential MEMC-Net model and (b) our proposed MEMC-Net model.}
		%
	}
	\label{fig:our_framework}
\end{figure}

\begin{table}[!t]
	\centering
    \footnotesize
	\caption{
		\textbf{CNN-based frame interpolation methods}.
	}
	\scriptsize
	\begin{tabular}{L{2.55cm}C{0.80cm}C{1.1cm}C{1.45cm}C{1.0cm}}
		\toprule
		& Optical flow &	Occlusion mask &	Interpolation coefficients & 	Kernel \quad size \\
		\midrule
	 Flow-based~\cite{liu2017video,xue2017video} &
		\checkmark &  \checkmark & 	fixed &	$2\times 2$ \\
	\midrule
		\vspace{0.1pt}
		Kernel-based~\cite{niklaus2017videoAdaConv,niklaus2017videoSepConv} &
		 --- & 	--- &  adaptive &$41 \times 41$, $51\times 51$ \\
		 \midrule
		 \vspace{0.5pt}
		\Ours{} (Ours)	& 	 \checkmark &\checkmark & adaptive& $4\times 4$ \\
		
		\bottomrule
	\end{tabular}
	\label{tab:compare_stoa}
	\vspace{-3mm}

\end{table}

\section{Motion Estimation and Motion Compensation Driven Neural Network }
\label{sec:Method}

In this section, we describe the design methodology of the proposed MEMC-Net framework, adaptive warping layer, and flow projection layer used in our model.

\begin{figure*}[!tb]
	\centering
		\renewcommand{\tabcolsep}{2pt} 
		\renewcommand{\arraystretch}{0.5} 
		\centering
	\includegraphics[width=0.98\textwidth]{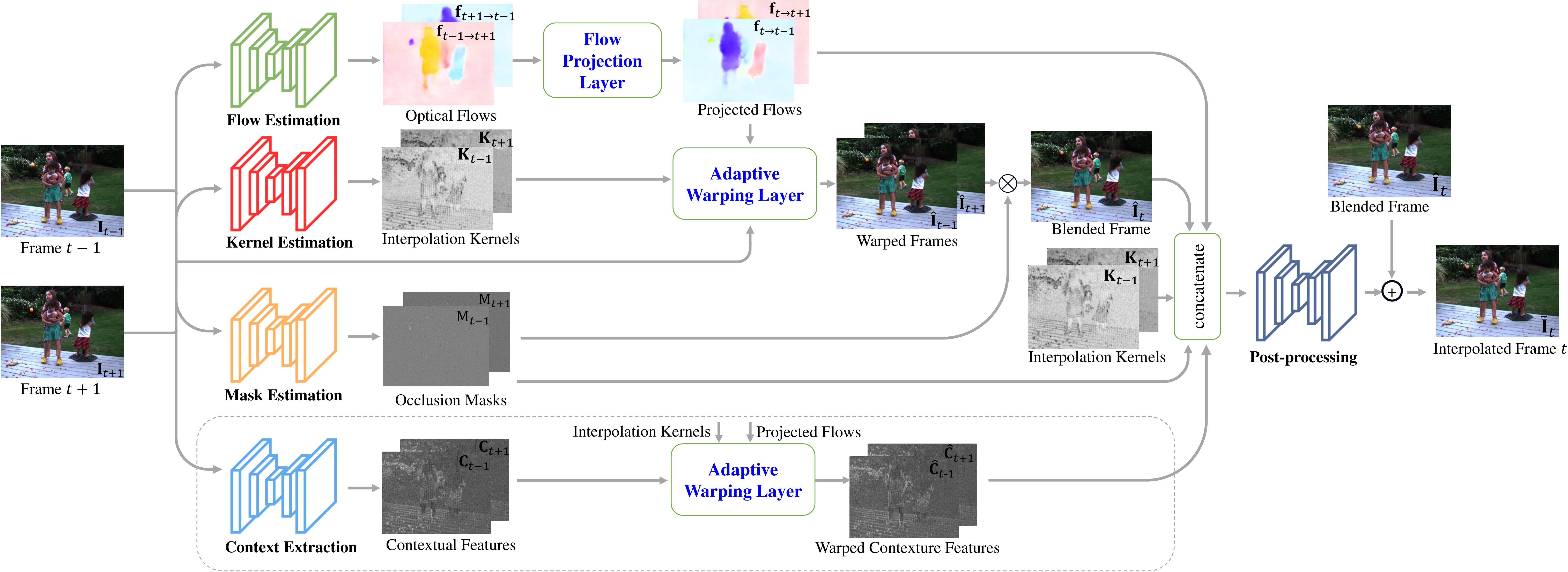}
		\vspace{-0mm}
	\caption{
		\textbf{Network architecture of the proposed \Ours{} and \Ours*.}
		The context extraction module and its generated contextual features and warped contextual features are for \Ours*.
	}
	\label{fig:net-arch}
\end{figure*}

\subsection{MEMC-Net Framework}
\label{sec:MEMC_arch}
Following the conventional MEMC-based and recent learning-based methods, there are different ways to design a MEMC-Net model for video frame interpolation. 
A straightforward method is to combine the motion estimation, motion compensation, and \lastNet{} sequentially.
That is, the reference frames are first aligned with the motion estimation, 
bilinear warping is applied to account for large motion, and
small convolutional kernels for the warped frames are estimated to synthesize a final frame.
{As in the conventional MEMC-based framework, a \lastNet network is also added to the sequential model to reduce the possible pixel outliers. 
}
\figref{our_framework}(a) illustrates this sequential model.
However, according to our experiments, the warped frames ($\hat{\mathbf{I}}_{t-1}$ and $\hat{\mathbf{I}}_{t+1}$) are usually of low quality due to the imperfect optical flow estimated by existing methods.
{
Consequently, the lateral kernel estimation, kernel convolution, and post-processing are not able to generate visually pleasing results from the corrupted frames.
}

In this paper, we develop a novel algorithm to simultaneously estimate the flow and compensation kernels with respect to the original reference frames.
This approach requires frame interpolation to be carried out within a warping layer based on both the flow and compensation kernel.
This new warping layer is expected to tightly couple the motion estimation and kernel estimation networks so that both networks can be optimized through the enormous video data.
\figref{our_framework}(b) shows the proposed framework for video frame interpolation.
%
%

We present the network architecture of the proposed MEMC-Net for the video frame interpolation in~\figref{net-arch}.
In this work, we propose a novel \emph{adaptive warping layer} to assemble the bilinear warping and kernel convolution in one single step.
The layer takes in the optical flow, interpolation kernel to warp the input frame pixels.
For the video frame interpolation task, since the intermediate frame is not originally available, 
we estimate the flow between the forward and backward reference frames, 
and then project it to simulate the flow between the intermediate and reference frames.
This operation is achieved by our proposed \emph{flow projection layer}.
The adaptive warping and the flow projection layers are the two major technical innovations of our algorithm.
{We summarize the benefits of the proposed layer from two aspects.
First, the conventional MEMC-based approaches rely on hand-crafted features (e.g., SIFT~\cite{brox2010large} for motion estimation or Gaussian-like weighting maps~\cite{wang2010motion} for motion compensation), while the proposed adaptive warping layer allows us to extract \textit{data-driven} features for joint motion estimation and motion compensation.
Therefore, the proposed model has a better \textit{generalization capability} to handle various scenarios for video frame interpolation and enhancement tasks.
Second, the adaptive warping layer tightly integrates two learning-based methodologies, namely the flow-based and kernel-based ones, and inherits their merits in that:\\
1) Compared to the flow-based methods~\cite{xue2017video,liu2017video} that rely on simple bilinear coefficients, our method is able to improve the \textit{interpolation accuracy} by using data-driven kernel coefficients. \\
2) Compared to the kernel-based approaches~\cite{niklaus2017videoAdaConv,niklaus2017videoSepConv}, our method obtains higher \textit{computational efficiency} by largely reducing the kernel size through pre-aligning the pixels with learned optical flows.
%
%
}
%
%

We present the forward inference and back-propagation details of the two novel layers in \secref{adaptive_warping} and \secref{flow_projection} respectively.
%
%
%
The pipeline of our method in~\figref{net-arch}, as well as the detailed network configuration of the used motion estimation, kernel estimation, mask estimation, and post-processing networks, are described in \secref{frame_interpolation}.
We will also introduce an additional context extraction network toward the enhanced \Ours* model.

\subsection{Adaptive Warping Layer}
\label{sec:adaptive_warping}
The proposed adaptive layer warps images or features based on the given optical flow and local convolutional kernels.
%
%

\begin{figure}[!t]
    \footnotesize
    \centering
    \begin{minipage}[b]{0.5\textwidth}
        \centering
        \includegraphics[width=0.95\linewidth]{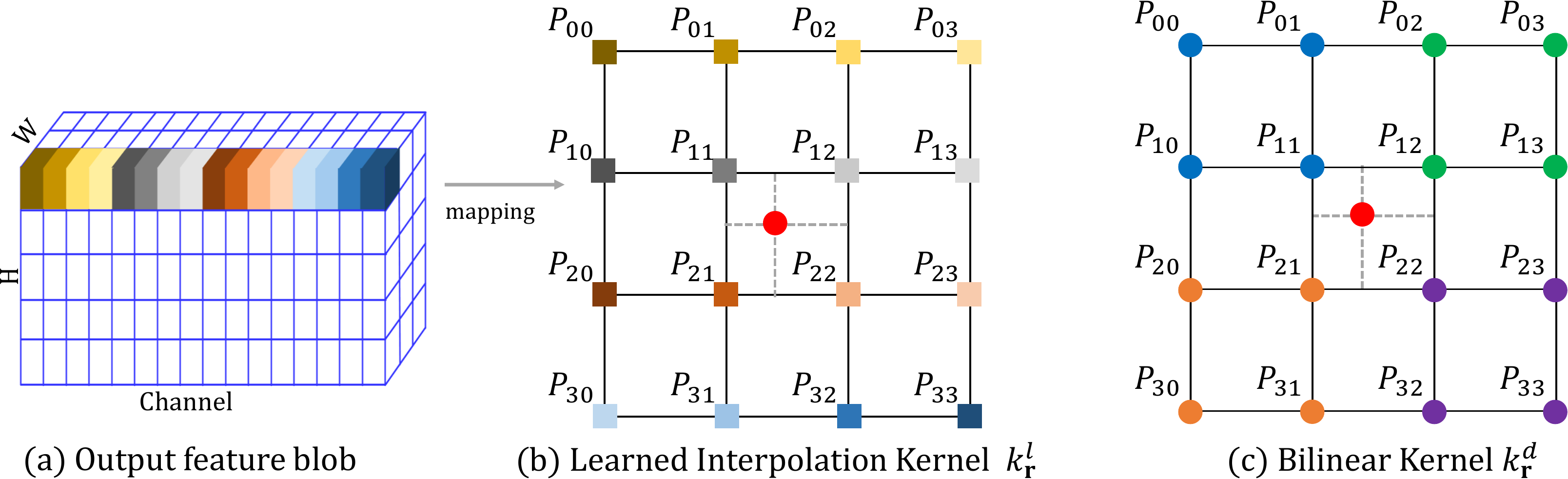}
        \caption{\textbf{Learned interpolation kernel $k^{l}_{\mathbf{r}}$ and bilinear kernel  $k^{d}_{\mathbf{r}}$.} The $k^{l}_{\mathbf{r}}$ is re-organized from the output feature blob generated by kernel estimation network.}
        \label{fig:kernel}
    \end{minipage}
    \vspace{-3mm}
\end{figure}

%
\Paragraph{Forward pass.}
Let $\mathbf{I}(\mathbf{x}): \mathbb{Z}^2\to\mathbb{R}^3$ denote the RGB image where $\mathbf{x}\in [1,H]\times[1,W]$, $\mathbf{f}(\mathbf{x})\coloneqq ( u(\mathbf{x}),v(\mathbf{x}))$ represent the optical flow field and $\mathbf{k}^{l} (\mathbf{x})= [k^{l}_{\mathbf{r}} (\mathbf{x})]_{H \times W}$ $(\mathbf{r} \in [-R+1, R]^2)$ indicate the interpolation kernel where $R$ is the kernel size.
The adaptive warping layer synthesizes an output image by:
\begin{equation}
\label{eq:adaptive_warping_forwad}
    \begin{split}
    \hat{\mathbf{I}}(\mathbf{x} ) =    \sum_{\mathbf{r}\in [-R+1,R]^2} k_{\mathbf{r}}(\mathbf{x}) \mathbf{I}(\mathbf{x} + \lfloor \mathbf{f}(\mathbf{x})\rfloor + \mathbf{r}), 
    \end{split}
\end{equation}
where the weight $k_{\mathbf{r}} = k^{l}_{\mathbf{r}} k^{d}_{\mathbf{r}}$ is determined by both the learned interpolation kernel $k^{l}_{\mathbf{r}}$ and bilinear coefficient $k^{d}_{\mathbf{r}}$.
{%
	We train a kernel estimation network to predict the weights for the interpolation kernels.
	For each 2D spatial location in the image grid $[1, W]\times[1,H]$, the kernel estimation network generates a $16$-channel feature vector.
	We then map the feature vector into a $4\times 4$ square matrix as the kernel coefficients for sampling the local patch.
	As shown in~\figref{kernel}(a) and (b), the colors on the feature vector and the patch pixels show the mapping of the $16$ channels.
	%
	%
	The red point in~\figref{kernel}(b) indicates the sub-pixel location shifted by the optical flow.
}

  	On the other hand, the bilinear coefficient (\figref{kernel}(c)) is defined by:
\begin{equation}
k^{d}_{\mathbf{r}} =
\left\{
\begin{aligned}
& [1-\theta(u)][1-\theta(v)],    &\mathbf{r}_u \leq 0 , \mathbf{r}_v \leq 0, \\
& \theta(u) [1-\theta(v)],        &\mathbf{r}_u > 0 , \mathbf{r}_v \leq 0, \\
& [1-\theta(u)] \theta(v),        &\mathbf{r}_u \leq 0 , \mathbf{r}_v > 0, \\
& \theta(u) \theta(v),             &\mathbf{r}_u > 0 , \mathbf{r}_v > 0, \\
\end{aligned}
\right .
\end{equation}
	where $\theta(u) = u- \lfloor u \rfloor$ denotes the fractional part of a float point number, and the subscript $u$, $v$ of the 2-D vector $\mathbf{r}$ represent the horizontal and vertical components, respectively.
	The bilinear coefficient allows the layer to back-propagate the gradients to the optical flow estimation network.
	%
	%
	In this case, we aim to compute a local interpolation kernel that combines the bilinear coefficients and the learned coefficients from the kernel prediction network.
	To apply the bilinear coefficients to kernels of any size, we first compute the bilinear coefficients for the nearest four neighbor pixels, i.e., $P_{11}$, $P_{12}$, $P_{21}$, and $P_{22}$, and then replicate the coefficients to the pixels at the same corner.
	Therefore, the pixels with the same color in~\figref{kernel}(c) have the same bilinear coefficient.
	Finally, we multiply the bilinear coefficients with the learned kernel coefficients as our local adaptive kernels.

%
%

\Paragraph{Backward pass.}
We compute the gradient with respect to the optical flow and interpolation kernels, respectively.
The derivative with respect to the optical flow field $\mathbf{f}$ is computed by (using the horizontal component $u$ for example):
\begin{equation}
\begin{split}
\frac{\partial{\hat{\mathbf{I}}(\mathbf{x})}}{\partial{u}(\mathbf{x})} 
= \sum_{\mathbf{r}} k_{\mathbf{r}}^{l}(\mathbf{x}) \cdot  \mathbf{I} \left( \mathbf{x} + \lfloor \mathbf{f}(\mathbf{x})\rfloor + \mathbf{r} \right) \cdot \frac{\partial k^{d}_{\mathbf{r}}}{\partial u}
\end{split}
,
\end{equation}
where 
\begin{equation}
\frac{\partial k^{d}_{\mathbf{r}}}{\partial u} =
\left\{
\begin{aligned}
& -[1-\theta(v)],&\mathbf{r}_u \leq 0 , \mathbf{r}_v \leq 0, \\
& [1-\theta(v)],&\mathbf{r}_u > 0 , \mathbf{r}_v \leq 0, \\
& -\theta(v),    &\mathbf{r}_u \leq 0 , \mathbf{r}_v > 0, \\
& \theta(v),     &\mathbf{r}_u > 0 , \mathbf{r}_v > 0. \\
\end{aligned}
\right .
\end{equation}
The derivative with respect to the vertical component $v$ can be derived in a similar way.

The derivative with respect to the interpolation kernel $k^{l}_{\mathbf{r}}$ is:
\begin{equation}
\frac{\partial{\hat{\mathbf{I}}}}{\partial k^{l}_{\mathbf{r}}(\mathbf{x})} =  k^{d}_{\mathbf{r}}(\mathbf{x}) \cdot \mathbf{I} \left(\mathbf{x} + \lfloor \mathbf{f}(\mathbf{x})\rfloor + \mathbf{r} \right).
\end{equation}

The integration with the spatially-varying kernels alleviates the limitation of 
bilinear interpolation to synthesize pixel values from a broader neighborhood.
In addition, this approach facilitates the warping layer to perform more robustly to inaccurate optical flow and better account for occlusion.

\subsection{Flow Projection Layer}
\label{sec:flow_projection}
As the intermediate frame is not available, we transform the flow between the forward and backward reference frames and then project it to simulate the flow between the intermediate frame and the reference frames. 
Let $\mathbf{f}_{t \to t-1}(\mathbf{x})$ be the motion vector field of frame $\mathbf{I}_{t}$ to $\mathbf{I}_{t-1}$.
Similarly, $\mathbf{f}_{t-1 \to t+1}(\mathbf{y})$ represents the motion vector field of frame $\mathbf{I}_{t-1}$ to $\mathbf{I}_{t+1}$.
Note that we use $\mathbf{y}$ to index the 2-D coordinate at time step $t-1$, as distinguished to $\mathbf{x}$ at $t$.
Our flow projection layer is designed to transform an estimated flow $\mathbf{f}_{t-1 \to t+1}(\mathbf{y})$ to $\mathbf{f}_{t \to t-1}(\mathbf{x})$.
%
Here we assume that the local motion between consecutive frames is linear and invert the flow between $\mathbf{I}_{t-1}$ and $\mathbf{I}_{t+1}$ to approximate the intermediate flow fields.
%
%

As there may exist multiple flow vectors projected to the same location in the intermediate frame, we average all the projected flow vectors at the same location.
On the other hand, there may exist holes where no flow is projected.
Thus, we use the outside-in strategy~\cite{baker2011database} to fill-in these holes in the intermediate frame.
We denote the set of flow vectors mapped to location $\mathbf{x}$ of time step $t$ by $\mathcal{S}(\mathbf{x}) := \{\mathbf{y}: \mathrm{round}\Big(\mathbf{y}+\mathbf{f}_{t-1 \to t+1}(\mathbf{y})/2 \Big) = \mathbf{x}, \forall\ \mathbf{y} \in [1,H]\times[1,W]  \}$ and denote the 4-directional nearest  available flow vectors of a hole by $\mathcal{N}(\mathbf{x}) := \{\mathbf{x}^{\prime}: |\mathcal{S}(\mathbf{x}^{\prime})| > 0 \}$.  
The forward pass of the proposed projection layer is defined by:
\begin{equation}
\label{eq:flow_project_forward}
\mathbf{f}_{t \to t-1} (\mathbf{x}) = \left\{
\begin{aligned}
& \frac{-1}{|\mathcal{S}(\mathbf{x})|}\sum_{\mathbf{y} \in \mathcal{S}(\mathbf{x})} \frac{\mathbf{f}_{t-1 \to t+1} (\mathbf{y})}{2},&\mbox{if \ } |\mathcal{S}(\mathbf{x})| > 0 , \\
& \frac{1}{|\mathcal{N}(\mathbf{x})|}\sum_{\mathbf{x}^{\prime} \in \mathcal{N}(\mathbf{x})}\mathbf{f}_{t \to t-1} (\mathbf{x}^{\prime}),&\mbox{if \ } |\mathcal{S}(\mathbf{x})| = 0.   \\
\end{aligned}
\right .
\end{equation}
The backward pass computes the derivative with respect to the input optical flow $ \mathbf{f}_{t-1 \to t+1}(\mathbf{y})$:
\begin{equation}
\frac{\partial \mathbf{f}_{t \to t-1} (\mathbf{x})\ \  }{\partial \mathbf{f}_{t-1 \to t+1} (\mathbf{y})} = \left\{
\begin{aligned}
& \frac{-1}{2 |\mathcal{S}(\mathbf{x})|}, &\mbox{for \ }\mathbf{y} \in \mathcal{S}(\mathbf{x})\ \mbox{if \ }|\mathcal{S}(\mathbf{x})| > 0 , \\
& 0, &\mbox{for \ }\mathbf{y} \notin \mathcal{S}(\mathbf{x})\ \mbox{or\ } |\mathcal{S}(\mathbf{x})| = 0.   \\
\end{aligned}
\right .
\end{equation}

\begin{figure}[!t]
	\centering
	\renewcommand{\tabcolsep}{2pt} 
	\renewcommand{\arraystretch}{0.5} 
	\centering
	\includegraphics[width=0.24\textwidth]{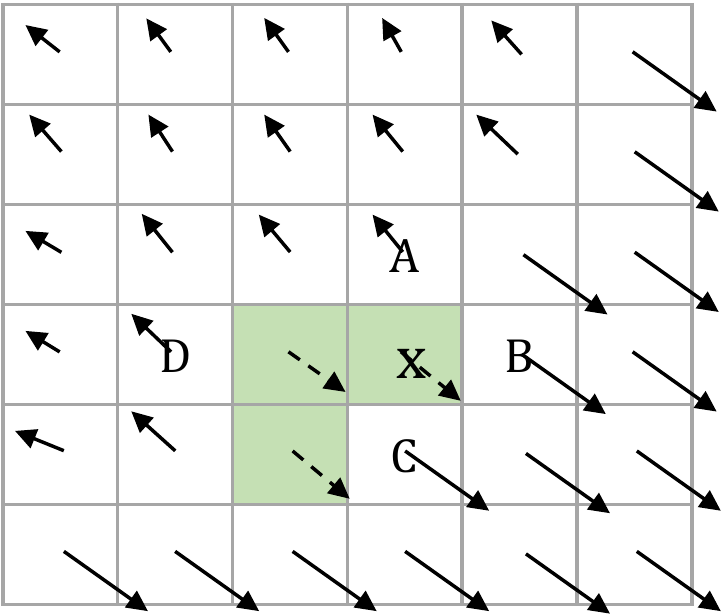}
	\vspace{-2mm}
	\caption{
		\textbf{Outside-in strategy for filling the flow holes.}
		The green regions indicate a hole, where the flow vectors are approximated by the average of 4-directional available flow vectors from the non-hole regions.
	}
	\label{fig:outside-in}
\end{figure}

\begin{figure}[!t]
	\footnotesize
	\centering
	\renewcommand{\tabcolsep}{1pt} 
	\renewcommand{\arraystretch}{1} 
	\begin{center}
		\begin{tabular}{cccc}

			\multicolumn{1}{c}{\includegraphics[width=0.24\linewidth]{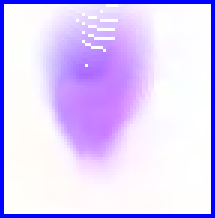}} &
			\multicolumn{1}{c}{\includegraphics[width=0.24\linewidth]{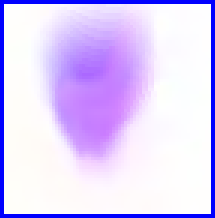}}&
			\multicolumn{1}{c}{\includegraphics[width=0.24\linewidth]{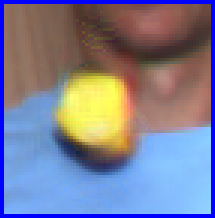}} &
			\multicolumn{1}{c}{\includegraphics[width=0.24\linewidth]{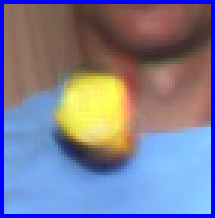}}
			\\
			
			(a) zero filling & (b) outside-in &
	(c) zero filling  & (d) outside-in  \\
	&&\ \ (IE = 2.76)&\ \ (IE = 2.66) \\
%
%
%
%
%
%
%
%
		\end{tabular}             
	\end{center}
	\vspace{-2mm}
	\caption{\textbf{Effectiveness of the used outside-in strategy for hole filling.}
		(a) and (b) are the flow maps by zero filling and outside-in strategy.
		(c) and (d) are the generated frames by them. Less artifact is generated by the outside-in hole filling strategy. 
		IE is short for interpolation error. The lower, the better.
		}
	\label{fig:outside-in-example} 
\end{figure}

\modify{We use a graph to illustrate the outside-in strategy in~\figref{outside-in}.
We use a \textit{soft} blending way in the proposed flow projection layer by averaging the 4-directional available flow vectors from the neighboring non-hole regions.
The spatial position at $\mathrm{X}$ has its 4-directional non-hole neighbors $\mathrm{A}$, $\mathrm{B}$, $\mathrm{C}$, and $\mathrm{D}$.
Therefore, the flow vector $\mathbf{f}_\mathrm{X}$ is approximated by $\mathbf{f}_\mathrm{X} = (\mathbf{f}_\mathrm{A} + \mathbf{f}_\mathrm{B} + \mathbf{f}_\mathrm{C} + \mathbf{f}_\mathbf{D})/4$.
%
%
An alternative is to fill in the flow holes with zero vectors, which is only suitable for stationary objects.
We show an example in~\figref{outside-in-example} to compare the two strategies.
The outside-in strategy can reduce interpolation artifacts significantly.}

\section{Video Frame Interpolation}
\label{sec:frame_interpolation}
%
{We provide an overview of the proposed MEMC-Net in~\figref{net-arch} and describe the detailed architecture design of each component below.}

\Paragraph{Motion estimation.} 
{Given two input frames $\mathbf{I}_{t-1}$ and $\mathbf{I}_{t+1}$, we first estimate the forward flow $\mathbf{f}_{t-1 \rightarrow t+1}$ and backward flow $\mathbf{f}_{t+1 \rightarrow t-1}$ by passing $\mathbf{I}_{t-1}$ and $\mathbf{I}_{t+1}$ into 
the flow estimation network twice with a reverse order.
In this work, we use the FlowNetS~\cite{dosovitskiy2015flownet} model for optical flow estimation.
Then we use the proposed flow projection layer as described by~Eq.(6) to project the forward flow $\mathbf{f}_{t-1 \rightarrow t+1}$ and backward flow $\mathbf{f}_{t+1 \rightarrow t-1}$ into $\mathbf{f}_{t \rightarrow t-1}$ and $\mathbf{f}_{t \rightarrow t+1}$ for the intermediate frame, respectively.}

\Paragraph{Kernel estimation.}
{We use the U-Net~\cite{ronneberger2015u} as our kernel estimation network, which has an encoder with five max-pooling layers, a decoder with five un-pooling layers, and skip connections from the encoder to the decoder.
The kernel prediction network takes two video frames as input and generates $R^2$ coefficient maps, denoted by $\mathbf{K}_{t-1}$ and $\mathbf{K}_{t+1}$.
We then reshape the coefficient maps to $R \times R$ convolutional kernels for each pixel, as shown in~\figref{kernel}(b).
Two pairs of intermediate flow and the kernel coefficients, \{$\mathbf{f}_{t \rightarrow t-1}$, $\mathbf{K}_{t-1}$\} and \{$\mathbf{f}_{t \rightarrow t+1}$, $\mathbf{K}_{t+1}$\} are then fed into the proposed adaptive warping layer to warp the input frames by Eq. (1) and generate two warped frames $\hat{\mathbf{I}}_{t-1}$ and $\hat{\mathbf{I}}_{t+1}$.}

\Paragraph{Mask estimation.} 
%
%
%
{Due to the depth variation and relative motion of objects, there are occluded pixels between the two reference frames.
To select valid pixels from the two warped reference frames, we learn a mask estimation network to predict the occlusion masks.
%
%
The mask estimation network has the same U-Net architecture as our kernel estimation network, but the last convolutional layer outputs a 2-channel feature map as the occlusion masks $\mathrm{M}_{t-1}$ and $\mathrm{M}_{t+1}$.
The blended frame is generated by:
\begin{equation}
\hat{\mathbf{I}}_t = \mathrm{M}_{t-1} \otimes \hat{\mathbf{I}}_{t-1} + \mathrm{M}_{t+1} \otimes \hat{\mathbf{I}}_{t+1}\,,
\end{equation}
where $\otimes$ denotes the channel-wise multiplication operation.}

\begin{figure}[!t]

	\centering	 
	\begin{minipage}{0.7\linewidth}
		\centering{\includegraphics[width= 0.8\textwidth]{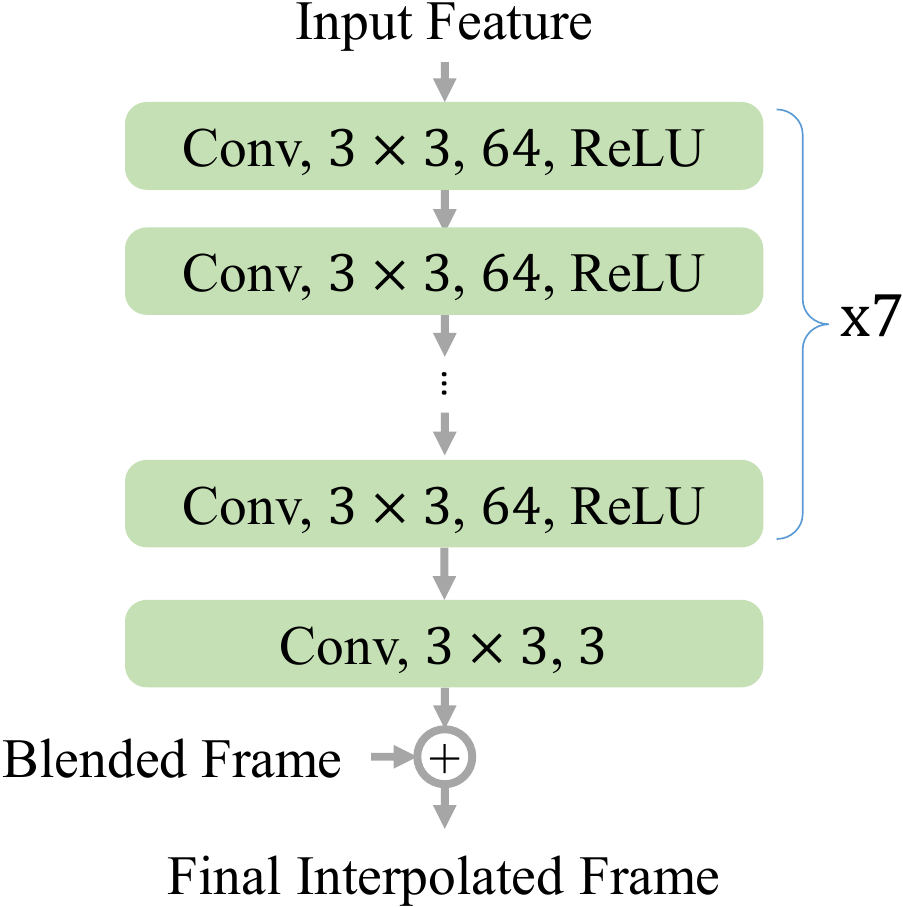}}
	\end{minipage}

	\vspace{-5pt}
	\caption{\textbf{Proposed \lastNet network.}
	}
	\label{fig:post-processing}
\end{figure}

\Paragraph{Context extraction.} 
%
%
%
{We also use the contextual information~\cite{niklaus2018context} in the \lastNet module to better deal with occlusion.
We extract the \textit{conv1} features of the input reference frames from a pre-trained ResNet18~\cite{he2016deep} as the contextual maps.
The contextual maps are then warped by the optical flow and the interpolation kernels via the adaptive warping layer.
The warped contextual maps, denoted as $\hat{\mathbf{C}}_{t-1}$ and $\hat{\mathbf{C}}_{t+1}$, are fed as inputs to the following \lastNet.}

\begin{figure}[!t]
\footnotesize
\centering
\renewcommand{\tabcolsep}{1pt} 
\renewcommand{\arraystretch}{1} 
\begin{center}
\begin{tabular}{lrlrlr}

	\multicolumn{2}{c}{\includegraphics[width=0.33\linewidth]{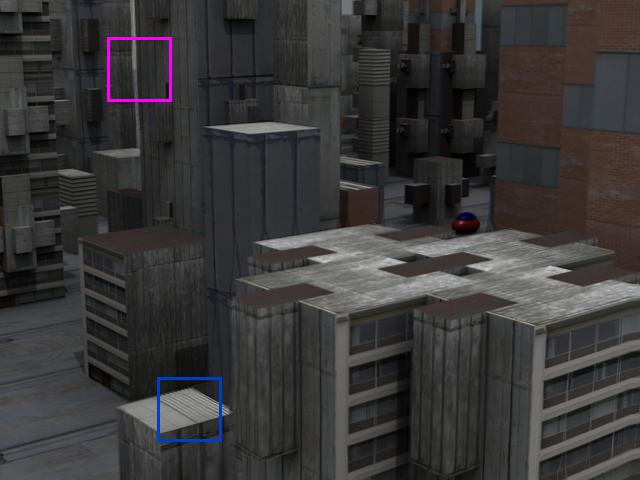}} &
	\multicolumn{2}{c}{\includegraphics[width=0.33\linewidth]{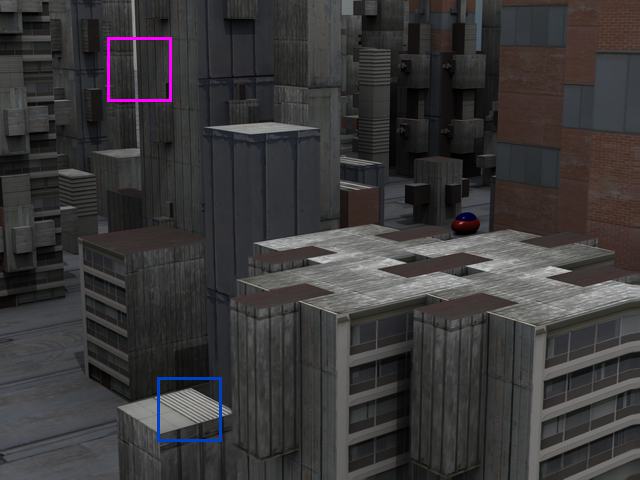}}&	
	\multicolumn{2}{c}{\includegraphics[width=0.33\linewidth]{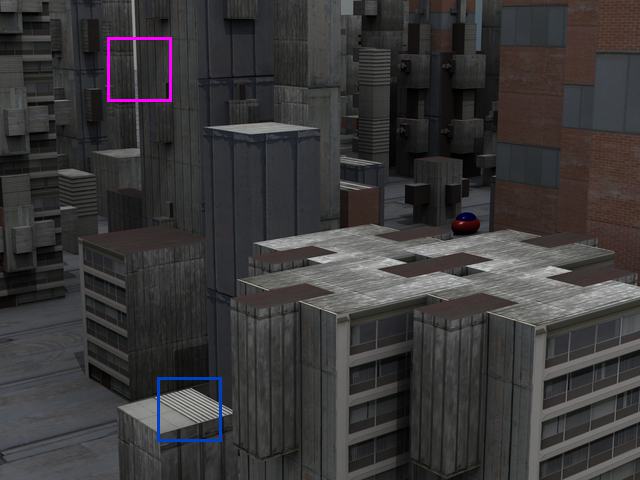}}		\\

	\includegraphics[width=0.162\linewidth]{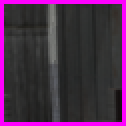} &
	\includegraphics[width=0.162\linewidth]{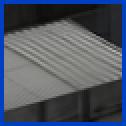} &

\includegraphics[width=0.162\linewidth]{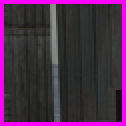} &
	\includegraphics[width=0.162\linewidth]{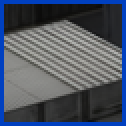} & 
	
	\includegraphics[width=0.162\linewidth]{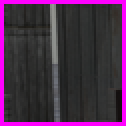} &
	\includegraphics[width=0.162\linewidth]{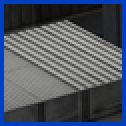}   \\
	
\multicolumn{2}{c}{Before post-processing}  & \multicolumn{2}{c}{After post-processing} & 	\multicolumn{2}{c}{Ground-truth} \\		
	
\end{tabular} 			
\end{center}
\vspace{-5pt}
\caption{\textbf{Effectiveness of the proposed post-processing network.}}
\label{fig:post-processing_effect} 
\end{figure}

\Paragraph{\LastNet.} 
{Since the blended image $\hat{\mathbf{I}}_t$ usually contains artifacts caused by inaccurate flow estimation or masks, we introduce a \lastNet network to improve the visual quality.
%
%
The \lastNet module takes as input the blended frame $\hat{\mathbf{I}}_t$, estimated flows $\mathbf{f}_{t \rightarrow t+1}$ and $\mathbf{f}_{t \rightarrow t-1}$, coefficient maps of the interpolation kernels $\mathbf{K}_{t-1}$ and $\mathbf{K}_{t+1}$, occlusion masks $\mathrm{M}_{t-1}$ and $\mathrm{M}_{t+1}$, and the warped context features $\hat{\mathbf{C}}_{t-1}$ and $\hat{\mathbf{C}}_{t+1}$.
Our post-processing network contains 8 convolutional layers as shown in~\figref{post-processing}.
Except for the last one, each convolutional layer has a filter size of $3\times 3$ with 64 output channels and is followed by a Rectified Linear Unit (ReLU).
The last convolutional layer outputs a 3-channel RGB image.
As the output and input of this module are highly similar (i.e., both are the interpolated frame at $t$), we enforce the network to output the \textit{residual} between the blended frame $\hat{\mathbf{I}}_t$ and the ground-truth frame.
%
%
%
Therefore, the \lastNet module learns to enhance the details and remove the artifacts in the blended frame.
We present an example in~\figref{post-processing_effect} to demonstrate the effect of the post-processing module.
The blurry edges and lines are sharpened by our method.
The proposed model generates $\tilde{\mathbf{I}}_t$ as the final interpolated frame.}

\Paragraph{MEMC-Net.}
{We provide two variants of the proposed model.
The first one does not use the contextual information, where we refer the model as \textit{\Ours}.
The second one includes the context information as the inputs to the post-processing module, where we refer the model as \textit{\Ours*}.}

\section{Implementation Details}
\label{sec:Implementation}
In this section, we discuss the implementation details including the loss function, datasets, and hyper-parameter settings of the proposed \Ours{} and \Ours*.

\Paragraph{Loss Function.} 
%
{
We use a robust loss function between the restored frames $\tilde{\mathbf{I}}_t$, $\hat{\mathbf{I}}_t$ and the corresponding ground truth frame $\mathbf{I}^{GT}_{t}$.
We also regularize the sum of two masks to be 1.0.
The combined loss function is given by:
\begin{equation}
\begin{split}
\mathcal{L} = &\sum_{\mathbf{x}} \Phi\left( \tilde{\mathbf{I}}_t - \mathbf{I}^{GT}_{t} \right) + \alpha \sum_{\mathbf{x}} \Phi\left( \hat{\mathbf{I}}_t - \mathbf{I}^{GT}_{t} \right) \\
& + \beta \sum_{\mathbf{x}}  \Phi\left(  \mathrm{M}_{t-1} + \mathrm{M}_{t+1} - 1.0 \right) \\
\end{split}
	\label{eq:loss}
\end{equation} 
where $\Phi(x) = \sqrt{x^2 +\epsilon^2}$ is the Charbonnier penalty function~\cite{charbonnier1994two} with  $\epsilon$ to be $1e-6$.
We empirically set $\alpha$ and $\beta$ to be $1e-3$ and $2e-3$ respectively.
}

%
%
\Paragraph{Datasets.}
We use the training set of the Vimeo90K dataset~\cite{xue2017video} to learn the 
proposed frame interpolation model.
There are 51,312 triplets, and each image is of $448 \times 256$ pixels.
During the training process, we use the data augmentation with random horizontal and vertical flipping as well as reversing the temporal order of input sequences.

\Paragraph{Hyper-parameter settings.}
We initialize the network parameters with the method of He~\etal~\cite{he2015delving}. 
We set the initial learning rate of the kernel prediction, mask estimation {and \lastNet} networks to be {0.001} while using a smaller learning rate of {0.00001} for fine-tuning the flow estimation network.
We decrease the learning rates by a factor of 0.2 if the validation loss does not decrease during 5 epochs.
We use a batch size of 4 and use the Adam~\cite{kingma2015adam} optimizer with $\beta_1$ of $0.9$ and $\beta_2$ of $0.999$ for training our model.
{In addition, we use a weight decay of $1e-6$.}
{The entire network is trained for 100 epochs.}
Except for the last output layer, the convolutional layers of the FlowNetS network are activated by the leaky ReLU~\cite{maas2013rectifier}, while those of the other three networks are activated by the ReLU~\cite{nair2010rectified} layer.
We use the batch normalization~\cite{ioffe2015batch} layer in the kernel prediction and mask estimation networks. 

The source code, trained model, and video frame interpolation results generated by all the evaluated methods are available on our project website: \url{https://sites.google.com/view/wenbobao/memc-net}. 

\begin{figure}[!t]
	\centering
		\renewcommand{\tabcolsep}{2pt} 
		\renewcommand{\arraystretch}{0.5} 
		\centering
			\begin{tabular}{cc}
	\includegraphics[width=0.257\textwidth]{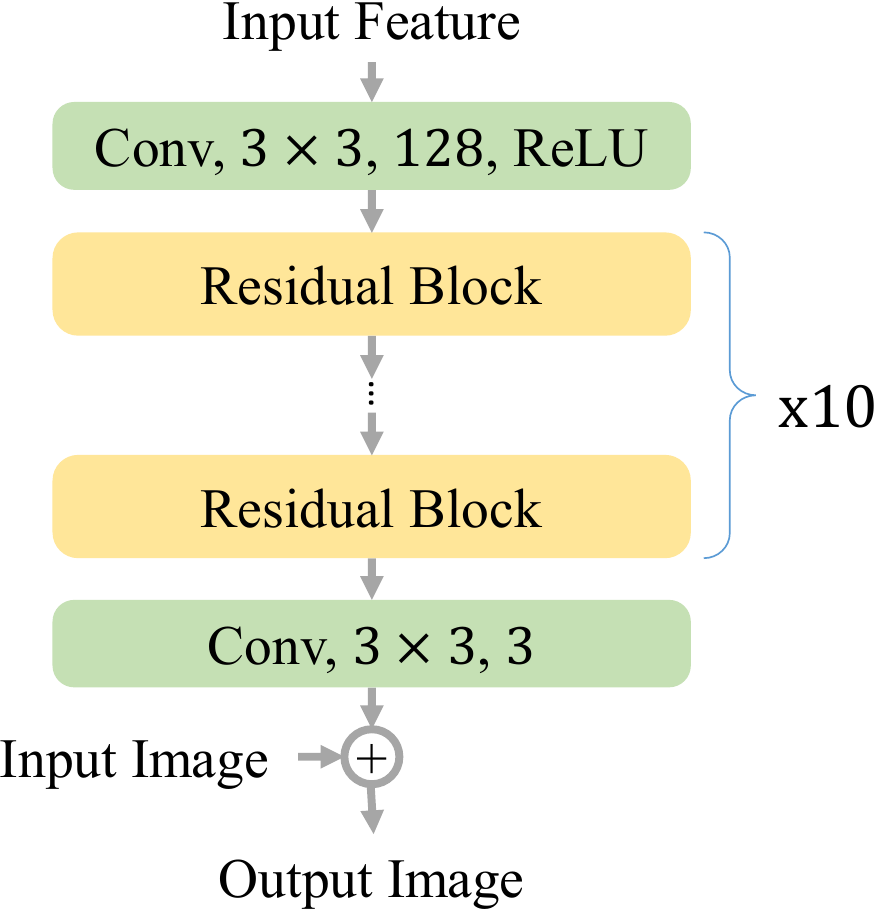}
				&
	\includegraphics[width=0.19\textwidth]{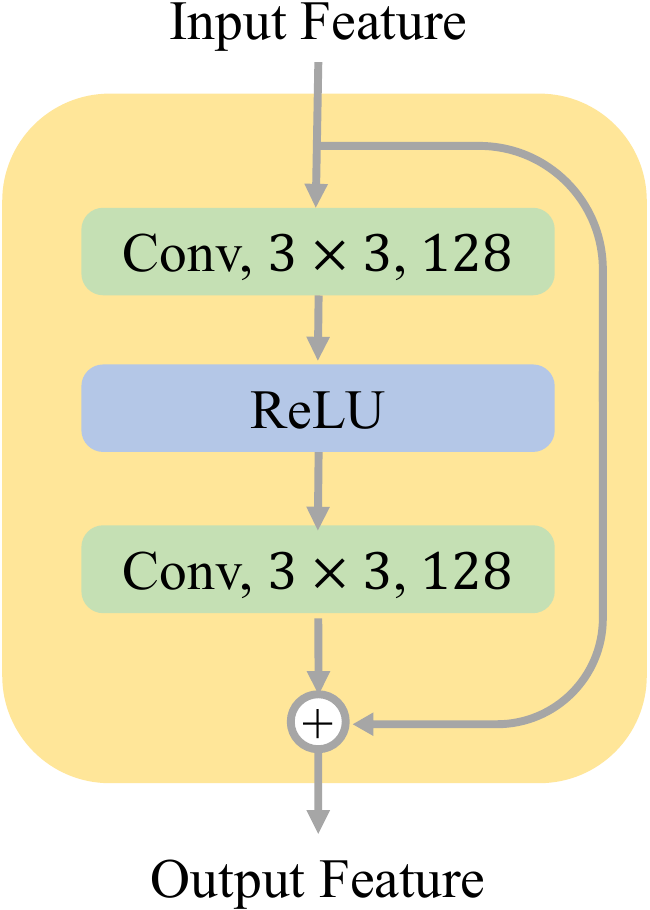}
				\\
				(a) Frame enhancement network & 
				(b) Residual block \\
			\end{tabular}
		\vspace{-0mm}
	\caption{
		\textbf{Network architecture for frame enhancement.}
	}
	\label{fig:frame_enhancement}
\end{figure}

\begin{figure}[!t]
	\centering
		\renewcommand{\tabcolsep}{2pt} 
		\renewcommand{\arraystretch}{0.5} 
		\centering
	\includegraphics[width=0.49\textwidth]{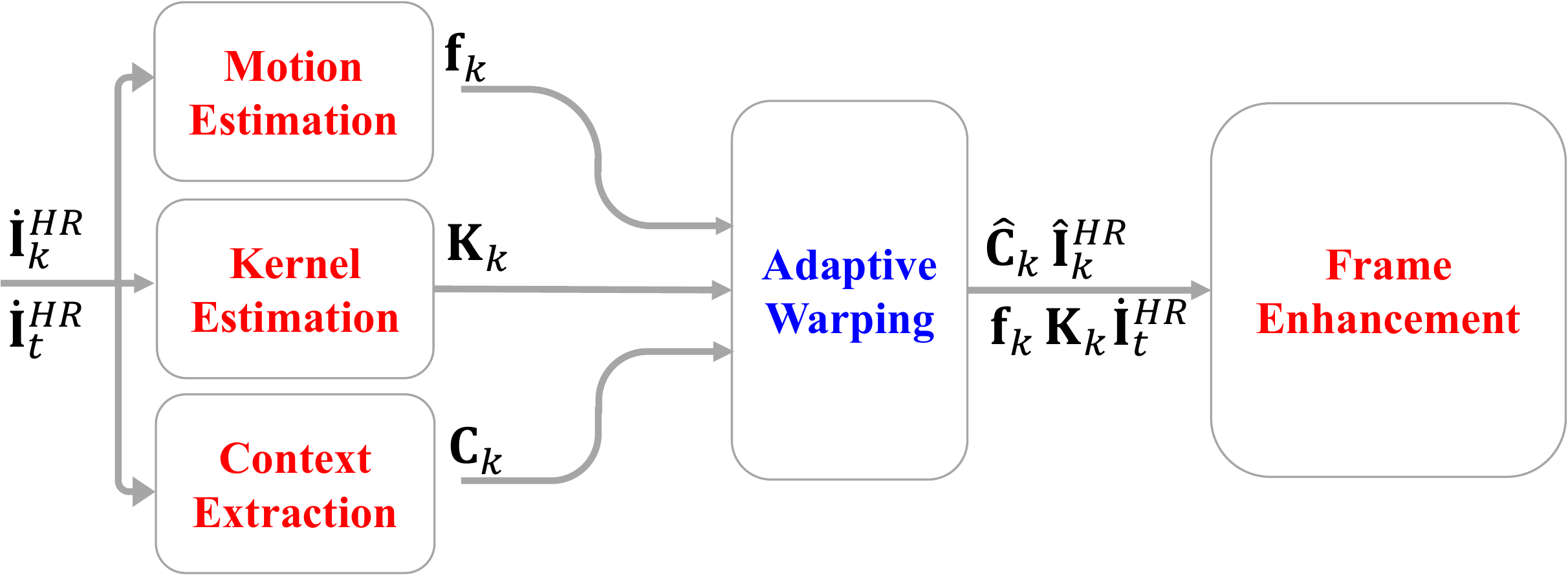}
		\vspace{-0mm}
	\caption{
		\textbf{Network architecture for video frame super-resolution.}
	}
	\label{fig:video_restoration_arch}
			\vspace{-5mm}
\end{figure}

\section{Video Frame Enhancement}
\label{sec:extensions}
In addition to video frame interpolation, we show that the proposed framework can be generalized to several video frame enhancement tasks, including video super-resolution, video denoising, and video deblocking.
In these tasks, multiple consecutive frames are used to extract useful texture cues to reduce the distortions like low-resolution, noise, blockiness, \etc.
Existing learning-based methods~\cite{liao2015video,xue2017video} often align the consecutive frames based on the estimated optical flow and bilinear interpolation.
However, as discussed in \secref{MEMC_arch} and \ref{sec:frame_interpolation}, 
the bilinear interpolation process may result in blurred pixels.
In contrast, the proposed adaptive warping layer is able to compensate for more accurate frame pixels.
Here we discuss how the proposed method can be extended for the video super-resolution problem.
Given $2L+1$ consecutive low-resolution frames $\{\mathbf{I}^{LR}_{k}\}^{t+L}_{k=t-L}$, our goal is to recover a high-resolution frame $\mathbf{\mathbf{I}}^{HR}_{t}$ at the time step $t$.
We first use bicubic interpolation to up-sample all the low-resolution frames to the target resolution, denoted by $\{\dot{\mathbf{I}}^{HR}_{k}\}^{t+L}_{k=t-L}$.
For each pair of $\dot{\mathbf{I}}^{HR}_{k} (k \neq t)$ and $\dot{\mathbf{I}}^{HR}_{t}$, we estimate the optical flow ${\mathbf{f}}_{k}(k \neq t)$ and compensation kernel ${\mathbf{K}}_{k}(k \neq t)$ via our flow estimation and kernel estimation networks.
Then, we use the proposed adaptive warping layer to warp all the neighboring frames to align with $\dot{\mathbf{I}}^{HR}_{t}$ at the time step $t$, denoted by $\hat{\mathbf{I}}^{HR}_{k} (k \neq t)$.
Alongside the frame pixels, we also extract and warp the context information from a pre-trained ResNet18~\cite{he2016deep} model.
Finally, all the generated motions ${\mathbf{f}}_{k}$, kernel ${\mathbf{K}}_{k}$, context $\hat{\mathbf{C}}_{k}$, warped frame $\hat{\mathbf{I}}^{HR}_{k}$ as well as 
up-sampled blurry frame $\dot{\mathbf{I}}^{HR}_{t}$ are fed into a frame enhancement network.
Our frame enhancement network has a similar architecture to the single-image 
super-resolution method, EDSR~\cite{lim2017enhanced}.

{The frame enhancement network is deeper than the post-processing network for video frame interpolation.
Since the input frames are heavily degraded by low resolution, noise, or blockiness, the frame enhancement network thus requires more complex architecture to restore high-quality results.
In the frame enhancement network shown in~\figref{frame_enhancement}(a), we first use one convolutional layer with a ReLU activation.
Then, we adopt 10 residual blocks, where each residual block contains two convolutional layers, one ReLU layer, and a skip connection as shown in~\figref{frame_enhancement}(b).
%
%
Finally, the last convolutional layer generates the residuals between the input and output images.
All the convolutional layers have 128 channels and a $3\times 3$ kernel size.
The final output frame of this network is denoted by $\tilde{\mathbf{I}}_{t}$.
We name the entire video super-resolution network as \Ours{\_SR} as shown in~\figref{video_restoration_arch}.
}

The model difference between the \Ours{\_SR} and \Ours{*} is twofold. 
First, the \Ours{\_SR} does not require the flow projection layer as we can directly estimate flow for the target frame.
Second, since each pixel of the target frame has a valid flow vector, we discard the mask estimation module in \Ours{\_SR}. 
%
We use the same network architecture as \Ours{\_SR} for video denoising and deblocking.
The extended model for denoising and deblocking are referred to as \Ours{\_DN} and \Ours{\_DB}, respectively.

{%
	For each of the three video enhancement tasks, we train our network on the corresponding training set from the Vimeo90K dataset~\cite{xue2017video}, namely Vimeo90K-SR, Vimeo90K-DN, and Vimeo90K-DB.  
	Each of the training sets consists of 91,701 7-frame sequences with an image resolution of $448 \times 256$ pixels.
	Note that the input images of the Vimeo90K-SR set are first downsampled to a resolution of $224 \times 128$ and then upsampled to $448 \times 256$ with the bicubic interpolation.
	In each training iteration, a batch contains one sequence with 7 consecutive frames.
	The learning rate is initialized to 0.0005 for the first 10 epochs and then is dropped to 0.0001 for the following 5 epochs.
	Similar to the video frame interpolation task, we initialized the parameters with the method of He~\etal~\cite{he2015delving} and use the Adamax optimizer~\cite{kingma2015adam} to update the parameters.
}

\begin{table*}[!t]
 
 \caption{\textbf{Analysis on flow-based methods.} The \first{red} numbers indicate the best performance. }
 \vspace{-8pt}
 \label{tab:flowbasedablation}
\footnotesize
\newcommand{\tabincell}[2]{\begin{tabular}{@{}#1@{}}#2\end{tabular}}
\newcolumntype{P}[1]{>{\centering\arraybackslash}p{#1}}

\centering
\begin{tabular}{m{2.0cm}  P{2.2cm}P{1.7cm}P{0.3cm}P{1.3cm} P{0.75cm} P{0.75cm}  P{0.75cm} P{0.75cm} P{1.8cm}  r}
\toprule
\multirow{2}{*}{Methods} & \multicolumn{4}{c}{Sub-networks}&  \multicolumn{2}{c}{UCF101~\cite{soomro2012ucf101}} &\multicolumn{2}{c}{Vimeo90K~\cite{xue2017video}} &Middlebury~\cite{baker2011database}
 &\multirow{2}{*}{\tabincell{c}{\#Param.}}   \\
\cmidrule(l{2pt}r{2pt}){2-5} 
\cmidrule(l{2pt}r{2pt}){6-7}
\cmidrule(l{2pt}r{2pt}){8-9} 
\cmidrule(l{2pt}r{2pt}){10-10}

&flow& kernel (size)& mask & \lastNetShort	&PSNR & SSIM 	&PSNR & SSIM	& 
IE \scriptsize{(\textit{oth.})}& \\
\midrule
DVF~\cite{liu2017video} & Enc-Dec & bilinear (2) &$\surd$ &$\times$ &34.12 & 0.9631 & 31.54 &  0.9462& 7.75 &1,604,547\\
ToFlow+Mask~\cite{xue2017video}&SPyNet &bilinear (2) & $\surd$ & $\surd$ & 34.58 & 0.9667  &33.73& 0.9682 & 2.51 &1,074,635\\
\midrule

\multirow{7}{1.8cm}{\Ours}
&FlowNetS &bilinear (2) & $\times$ & $\times$  &34.65  &0.9664   &32.73 &0.9606 & 2.81& 38,676,496 \\
&FlowNetS &bilinear (2) & $\times$ & $\surd$  & 34.70 & 0.9667  & 33.25& 0.9646 & 2.50& 38,922,323 \\
&FlowNetS &bilinear (2) & $\surd$ & $\times$  &34.69  & 0.9667  & 32.94& 0.9626&  2.72& 52,842,818  \\
&FlowNetS &bilinear (2) & $\surd$ & $\surd$  &34.76  &  0.9671 & 33.40 &0.9661 & 2.47& 53,088,645  \\
&FlowNetS &learned (4) & $\times$ & $\surd$  &34.77 & 0.9669&33.29 & 0.9663&2.42& 53,092,995\\
&FlowNetS &learned (4) & $\surd$ & $\times$  & {34.88} &0.9669 & 33.51 & 0.9670 & 2.37 & 67,013,490 \\

&FlowNetS (fixed) &learned (4) & $\surd$ & $\surd$  & {33.15} & {0.9632} &{32.05} &{0.9580} &{3.26} & {28,583,973}\\ 

&FlowNetS &learned (4) & $\surd$ & $\surd$  & \first{34.95} & \first{0.9679} &\first{34.02}& \first{0.9704} &\first{2.24} & 67,260,469  \\

\bottomrule
\end{tabular} 

\end{table*}
\begin{table*}[!t]
 
 \caption{\textbf{Analysis on kernel-based methods.}}
 \vspace{-8pt}
 \label{tab:kernelbasedablation}
\footnotesize
\newcommand{\tabincell}[2]{\begin{tabular}{@{}#1@{}}#2\end{tabular}}
\newcolumntype{P}[1]{>{\centering\arraybackslash}p{#1}}

\centering
\begin{tabular}{m{2.2cm}  P{1.0cm}P{1.7cm}P{0.6cm}P{1.8cm} P{0.75cm} P{0.75cm}  P{0.75cm} P{0.75cm} P{1.8cm}  r}
\toprule
\multirow{2}{*}{  Methods}& \multicolumn{4}{c}{Sub-networks} &  \multicolumn{2}{c}{UCF101~\cite{soomro2012ucf101}} &\multicolumn{2}{c}{Vimeo90K~\cite{xue2017video}} &Middlebury~\cite{baker2011database}
 &\multirow{2}{*}{\tabincell{c}{\#Param.}}   \\
\cmidrule(l{2pt}r{2pt}){2-5} 
\cmidrule(l{2pt}r{2pt}){6-7}
\cmidrule(l{2pt}r{2pt}){8-9} 
\cmidrule(l{2pt}r{2pt}){10-10}

&flow& kernel (size)& mask & \lastNetShort&PSNR & SSIM 	&PSNR & SSIM	& 
IE \scriptsize{(\textit{oth.})}& \\
\midrule
SepConv-$L_f$~\cite{niklaus2017videoSepConv} & $\times$  &learned (51) & $\times$ & $\times$ & 34.69 & 0.9655  & 33.45& 0.9674& 2.44 &21,675,452\\
SepConv-$L_1$~\cite{niklaus2017videoSepConv} &$\times$  &learned (51) & $\times$ & $\times$ & 34.78 & 0.9669  &33.79&  {0.9702}& {2.27} & 21,675,452 \\
\midrule
\multirow{3}{*}{\Ours}
&$\times$ &learned (4) & $\times$ & $\times$ & 34.89 &\first{0.9682} &32.73& 0.9581 &2.74& 14,710,415\\
&$\times$ &learned (4) & $\times$ & $\surd$ &  {34.97} &\first{0.9682}&33.31& 0.9633& 2.57& 14,720,783\\
&FlowNetS &learned (4) & $\surd$ & $\surd$  &\first{34.95} & {0.9679} &\first{34.02}& \first{0.9704} &\first{2.24}  & 67,260,469  \\

\bottomrule
\end{tabular} 
\vspace{-8pt}
\end{table*}

\begin{table}[!t]
	\begin{minipage}{1.0\linewidth}
		\caption{\textbf{Evaluation on models with fewer model parameters.} M.B. stands for Middlebury.}
		\vspace{-8pt}
		\label{tab:smallmodel}
		\footnotesize
		\setlength{\belowcaptionskip}{-0.8cm}
		\newcommand{\tabincell}[2]{\begin{tabular}{@{}#1@{}}#2\end{tabular}}
		\newcolumntype{P}[1]{>{\centering\arraybackslash}p{#1}}
		
		\centering
		\begin{tabular}{m{1.75cm} P{0.40cm} P{0.40cm}  P{0.40cm} P{0.40cm} P{0.9cm}  r}
			\toprule
			\multirow{2}{*}{Methods} & \multicolumn{2}{c}{UCF101~\cite{soomro2012ucf101}} &\multicolumn{2}{c}{Vimeo90K~\cite{xue2017video}} &M.B.~\cite{baker2011database}
			&\multirow{2}{*}{\#Param.}   \\
			\cmidrule(l{2pt}r{2pt}){2-3} 
			\cmidrule(l{2pt}r{2pt}){4-5}
			\cmidrule(l{2pt}r{2pt}){6-6}
			%
			%
			&PSNR & SSIM 	&PSNR & SSIM	& 
			IE~\scriptsize{(\textit{oth.})} & \\
			\midrule
			\Ours\_s & 34.83 & {0.9676} &{33.97} &{0.9721} &{2.44} & \first{7.2M}\\ 
			\Ours  & {34.95} & {0.9679} &{34.02}& {0.9704} & {2.24}& 67.2M \\

\Ours* & \first{35.01} & \first{0.9683} &\first{34.40}& \first{0.9742} &\first{2.10} &    70.3M \\ 
			\bottomrule
		\end{tabular} 
	\end{minipage}
	\vspace{-5pt}
	
\end{table}

\begin{table}[!t]


\caption{\textbf{Runtime of frame interpolation methods (seconds).}}
\vspace{-8pt}
\label{tab:running-speed}
\footnotesize
\setlength{\belowcaptionskip}{-0.8cm}
\newcommand{\tabincell}[2]{\begin{tabular}{@{}#1@{}}#2\end{tabular}}
\newcolumntype{P}[1]{>{\centering\arraybackslash}p{#1}}

\centering
\begin{tabular}{m{1.9cm}  P{1.6cm} P{1.7cm} P{1.8cm}   }
	\toprule
	Methods & $640 \times 480$p& $1280 \times 720$p &$1920 \times 1080$p 	 \\
	\midrule
	AdaConv~\cite{niklaus2017videoAdaConv}  & 2.80& --- &  ---\\
	ToFlow~\cite{xue2017video} & 0.43& 1.01& 1.90 \\
	SepConv~\cite{niklaus2017videoSepConv}  &0.20  & 0.50  & 0.90  \\
	\Ours\_s & 0.13 & 0.33& 0.67\\
	\Ours   & \first{0.06} & \first{0.20} & \first{0.41}\\
	\Ours*   & {0.12} & {0.36} & {0.64}\\	
	\bottomrule
\end{tabular} 
\vspace{-8pt}

\end{table}

\begin{table}[t]
	\caption{\textbf{Runtime of the proposed models (seconds).}  We evaluate these models on $640 \times 480$ videos.}
	\vspace{-8pt}
	\label{tab:detailtime}
	\footnotesize
	\setlength{\belowcaptionskip}{-0.8cm}
	\newcommand{\tabincell}[2]{\begin{tabular}{@{}#1@{}}#2\end{tabular}}
	\newcolumntype{P}[1]{>{\centering\arraybackslash}p{#1}}
	
	\centering
    	\begin{tabular}{m{1.8cm}  P{0.5cm} P{0.5cm} P{0.5cm} P{0.6cm} P{1.3cm} P{0.6cm}  }
		\toprule 
		Networks & flow   & kernel   & mask & context   &  \lastNetShort  & Total	 \\
		\midrule 
		\Ours\_s & 0.103 & 0.005& ---  & --- & 0.020& 0.13 \\
		\Ours   &  0.024&0.008&0.008&  ---  &0.020& 0.06\\
		\Ours*  &  0.024&0.008&0.008& 0.001&  0.080& 0.12\\		
		\bottomrule
	\end{tabular} 
	\vspace{-8pt}
\end{table}

\begin{table*}[!t]
	\caption{
		\textbf{Quantitative evaluation on UCF101, Vimeo90K, and Middlebury datasets.}
		The abbreviations \textit{oth.} and \textit{eval.} represent the \textsc{Other} and \textsc{Evaluation} sets in the Middlebury dataset.
		The runtime is evaluated on the $640\times 480$ sequences.
	}
	\label{tab:like-to-like}
	\footnotesize
	\newcommand{\tabincell}[2]{\begin{tabular}{@{}#1@{}}#2\end{tabular}}
	\newcolumntype{P}[1]{>{\centering\arraybackslash}p{#1}}
	
	\centering
	\begin{tabular}{m{2.5cm} P{1.5cm}P{1.5cm} P{1.1cm} P{1.1cm}  P{1.1cm} P{1.1cm}  P{1.4cm} P{1.4cm} }
		\toprule
		\multirow{2}{*}[-0.2em]{Methods} & \multirow{2}{*}[-0.2em]{\tabincell{c}{\#Parameters \\ (million)}} & \multirow{2}{*}[-0.2em]{\tabincell{c}{Runtime \\(seconds)}} &  \multicolumn{2}{c}{UCF101~\cite{soomro2012ucf101}} &\multicolumn{2}{c}{Vimeo90K~\cite{xue2017video}} &\multicolumn{2}{c}{Middlebury~\cite{baker2011database}}  \\
		\cmidrule(l{2pt}r{2pt}){4-5}
		\cmidrule(l{2pt}r{2pt}){6-7}
		\cmidrule(l{2pt}r{2pt}){8-9}
		&    & & PSNR & SSIM     &PSNR & SSIM    & 
		IE \scriptsize{(\textit{oth.})} & IE \scriptsize{(\textit{eval.})} \\
		\midrule
		ToFlow+Mask~\cite{xue2017video} & \first{1.07} &0.43 & 34.58 & 0.9667  &33.73& 0.9682 &2.51& --- \\
		
		SepConv-$L_1$~\cite{niklaus2017videoSepConv} & 21.6 &0.20 & 34.78 & 0.9669  &33.79& {0.9702}& {2.27} & 5.61  \\

		\Ours\_s & \second{7.20} &0.13&34.83 & {0.9676} &{33.97} &\second{0.9721} &{2.44} & --- \\ 

		\Ours& 67.2 & \first{0.06} & \second{34.95} & \second{0.9679} &\second{34.02}& {0.9704} &\second{2.24} &  \second{5.35}  \\

		\Ours* & 70.3 & \second{0.12} & \first{35.01} & \first{0.9683} &\first{34.40}& \first{0.9742} &\first{2.10} &  \first{5.00}  \\
		
		\bottomrule
	\end{tabular} 
	
\end{table*}

\section{Experimental Results}
\label{sec:ER} 
In this section, we first analyze and discuss the contributions of each sub-module, processing speed, and model parameters.
We then present the experimental results on video frame interpolation 
and the other three video frame enhancement tasks.

\subsection{Analysis and Discussion}
\label{subsec:AS}
We first describe the evaluated datasets and conduct experiments to analyze the contribution of each component in the proposed model, especially on flow and kernel estimation. 

\subsubsection{Datasets}
We evaluate the proposed frame interpolation approach on a wide variety of video datasets.

\Paragraph{Middlebury.}
The Middlebury dataset~\cite{baker2011database} is widely used for evaluation of optical flow estimation, stereo image matching, and frame interpolation methods. 
There are 12 sequences in the \textsc{Other} set and 8 sequences in the \textsc{Evaluation} set with a resolution of $640 \times 480$ pixels.
We use the evaluation protocol to compute the \textit{Interpolation Error} (IE) and \textit{Normalized Interpolation Error} (NIE).

\Paragraph{UCF101.}
The UCF101 dataset~\cite{soomro2012ucf101} contains a large variety of human actions. 
%
We use 379 triplets from the UCF101 test set, where the image resolution is $256 \times 256$ of pixels.  

\Paragraph{Vimeo90K.}
Xue~\etal~\cite{xue2017video} develop a high-quality dataset with videos from Vimeo (\url{https://vimeo.com}).
There are 3,782 triplets for evaluation with the image resolution of $448 \times 256$ pixels.

\Paragraph{HD videos.}
In this work, we collect 7 HD (High Definition) videos from the Xiph website ({\url{https://media.xiph.org/video/derf/}), and 
	interpolate the first 50 even frames for each of the videos.
	We also evaluate on four short video clips from the Sintel dataset~\cite{butler2012naturalistic}, where each image is of $1280 \times 544$ pixels. 

\subsubsection{Ablation Studies}
\Paragraph{Flow-based analysis.}
We first construct a baseline model by using the optical flow estimation network and bilinear interpolation for warping images.
Similar to the ToFlow~\cite{xue2017video} method, the baseline model does not contain the kernel estimation, mask estimation, and \lastNet networks.
\tabref{flowbasedablation} shows the performance of this baseline model is similar to that by the ToFlow method.
We then include the mask estimation and \lastNet (abbreviated by \lastNetShort{} in \tabref{flowbasedablation}) networks, where both modules clearly contribute to the performance on all the test datasets.
By replacing the fixed bilinear kernel module with the proposed spatially-adaptive kernel estimation network, a significant performance gain can be achieved.
Our final model with all the components achieves the state-of-the-art performance on all three benchmark datasets.
\modify{The FlowNetS contains 38.6M parameters, which take 57.4\% of the parameters in our MEMC-Net model.
We conduct an experiment to use a fixed flow estimation network.
The performance of this variant drops a lot on all the datasets, as shown in the last two rows of~\tabref{flowbasedablation}.
Without fine-tuning the flow estimation network, the projected flow is a simple approximation of the flow from the intermediate frame to the reference frame, which is not able to synthesize high-quality intermediate frame.
}

\Paragraph{Kernel-based analysis.}
We conduct experiments to analyze the contribution of the learned interpolation kernels in the proposed method.
We train a baseline model by removing the optical flow estimation network and only learn $4 \times 4$ spatially-adaptive kernels for interpolation.
This baseline model is similar to the SepConv method~\cite{niklaus2017videoSepConv} but with a much smaller kernel.
In~\tabref{kernelbasedablation}, it is interesting to see that this baseline model already outperforms the SepConv method on the UCF101 dataset.
It is sufficient to use a $4 \times 4$ interpolation kernel for all videos as
the image resolution is low and object motion is small.
However, it does not perform well on the Vimeo90K and Middlebury datasets, which contain much larger motion.
By introducing the flow estimation network, the proposed method performs 
better than the evaluated models on the Vimeo90K and Middlebury datasets.
The results demonstrate the importance of integrating optical flow and learned interpolation kernels to deal with large motions in frame interpolation.

%
%
%
%
%
%
%

\Paragraph{Model parameters.}
Since modern mobile devices typically have limited memory size, we present a smaller model with fewer parameters but maintaining the same MEMC framework.
We first replace the FlowNetS~\cite{dosovitskiy2015flownet} with the SPyNet~\cite{ranjan2017optical} model, which reduces about $97\%$ parameters in the flow estimation network.
We then simplify the kernel prediction network by removing one convolutional layer before each max-pooling and un-pooling layer and discard the occlusion estimation network.
This reduced model, denoted by \Ours{\_s}, has only 7,204,367 trainable parameters, which is $89.3\%$ smaller than our full network.
We compare the performance of the full and small models in~\tabref{smallmodel}. 
As the SPyNet performs better than the FlowNetS~\cite{dosovitskiy2015flownet} on small motion~\cite{ranjan2017optical}, the performance of \Ours{\_s} is slightly better than our full model on the Vimeo90K dataset.
However, \Ours{\_s} does not perform as well on the Middlebury dataset as it contains large displacement.

\Paragraph{Execution speed.}
We evaluate the runtime of the proposed algorithm on an NVIDIA Titan X (Pascal) GPU.
We compare the execution time of the proposed method and the state-of-the-art algorithms on $640 \times 480$p, $1280 \times 720$p and $1920 \times 1080$p videos in~\tabref{running-speed}.
Our \Ours{} model can process $1920 \times 1080$p videos with the runtime of $0.41$ second per frame.
Moreover, when using four GPU cards to process a $1920 \times 1080$p videos in parallel by splitting input frames into $270\times240$ non-overlapped patches, our method is able to process $30$ frames per second.
We note that the small model \Ours\_s does not necessarily have better runtime performance as the SPyNet applies several convolutional layers on the input resolution (which results in larger feature maps and a higher computational cost).
On the other hand, the operations of the FlowNetS model are mostly applied on the $1/4$ or smaller resolution space.
The SPyNet uses $97\%$ fewer parameters but has $1.35$ times more FLOPs than the FlowNetS.
In~\tabref{detailtime}, we show the runtime of each component in the proposed models.
The small model can be used for memory-constrained devices while the full model is preferable for applications that require prompt response time.
The proposed \Ours* can be used for cases where the interpolation quality is of most importance. 

\modify{
	Our \Ours\_s consists of only 7.2M parameters but performs better than a larger SepConv model which contains 21.7M parameters.
	Furthermore, the amount of parameters is not the only factor to be considered.
	Although the ToFlow+Mask model uses fewer parameters, it runs slower than the proposed and SepConv methods.
	%
	%
	We present the evaluation results of these algorithms in~\tabref{like-to-like}.
	}

\begin{table}[!t]
	\begin{minipage}{1.0\linewidth}
		\caption{\textbf{Quantitative comparisons with the sequential model.} M.B. stands for Middlebury.}
		\vspace{-8pt}
		\label{tab:sequentialmodel}
		\footnotesize
		\setlength{\belowcaptionskip}{-0.8cm}
		\newcommand{\tabincell}[2]{\begin{tabular}{@{}#1@{}}#2\end{tabular}}
		\newcolumntype{P}[1]{>{\centering\arraybackslash}p{#1}}
		
		\centering
		\begin{tabular}{m{1.75cm} P{0.40cm} P{0.40cm}  P{0.40cm} P{0.40cm} P{0.9cm}  r}
			\toprule
			\multirow{2}{*}{Methods} & \multicolumn{2}{c}{UCF101~\cite{soomro2012ucf101}} &\multicolumn{2}{c}{Vimeo90K~\cite{xue2017video}} &M.B.~\cite{baker2011database}
			&\multirow{2}{*}{\#Param.}   \\
			\cmidrule(l{2pt}r{2pt}){2-3} 
			\cmidrule(l{2pt}r{2pt}){4-5}
			\cmidrule(l{2pt}r{2pt}){6-6}
			%
			%
			&PSNR & SSIM 	&PSNR & SSIM	& 
			IE~\scriptsize{(\textit{oth.})} & \\
			\midrule
			Sequential & {34.34} & {0.9652} &{32.94} &{0.9639} &{2.47} & {67.2M}\\ 
			\Ours  & \first{34.95} & \first{0.9679} &\first{34.02}& \first{0.9704} & \first{2.24}& 67.2M \\ 
						
			
			\bottomrule
		\end{tabular} 
	\end{minipage}
	\vspace{-5pt}
	
\end{table}

\begin{table*}[!th]
	\caption{
		\textbf{Quantitative results on the Middlebury} \textsc{Evaluation} \textbf{set.} 
		%
		The \first{{red}} numbers indicate that corresponding method takes the 1st place among all the evaluated algorithms.
		%
	}
	\label{tab:Middlebury}
	\footnotesize
	\setlength{\belowcaptionskip}{0.8cm}
	\newcommand{\tabincell}[2]{\begin{tabular}{@{}#1@{}}#2\end{tabular}}
	\newcolumntype{P}[1]{>{\centering\arraybackslash}p{#1}}
	
	\centering
	\begin{tabular}{m{2.15cm}  P{0.44cm}P{0.44cm} P{0.44cm}P{0.44cm} P{0.44cm}P{0.44cm} P{0.44cm}P{0.44cm} P{0.44cm}P{0.44cm} P{0.44cm}P{0.44cm} P{0.44cm}P{0.44cm} P{0.44cm}P{0.44cm}   P{0.44cm}P{0.44cm} }
		\toprule
		\multirow{2}{*}[-0.2em]{Methods}& \multicolumn{2}{c}{Mequon} & \multicolumn{2}{c}{Schefflera} & \multicolumn{2}{c}{Urban} & \multicolumn{2}{c}{Teddy} & \multicolumn{2}{c}{Backyard} & \multicolumn{2}{c}{Basketball} & \multicolumn{2}{c}{Dumptruck} & \multicolumn{2}{c}{Evergreen} & \multicolumn{2}{c}{{Average}}\\
		\cmidrule(l{2pt}r{2pt}){2-3}
		\cmidrule(l{2pt}r{2pt}){4-5}
		\cmidrule(l{2pt}r{2pt}){6-7}
		\cmidrule(l{2pt}r{2pt}){8-9}
		\cmidrule(l{2pt}r{2pt}){10-11}
		\cmidrule(l{2pt}r{2pt}){12-13}
		\cmidrule(l{2pt}r{2pt}){14-15}
		\cmidrule(l{2pt}r{2pt}){16-17}
		\cmidrule(l{2pt}r{2pt}){18-19}
		&IE & NIE&IE & NIE&IE & NIE&IE & NIE&IE & NIE&IE & NIE&IE & NIE&IE & NIE&IE & NIE\\
		\midrule
		
		EpicFlow~\cite{revaud2015epicflow} 		    &3.17	&0.62 	&3.79    &0.70      &4.28        &1.06      &6.37     &1.09 	 &11.2      &1.18 &6.23   &1.10    &8.11 &1.00    	&8.76      &1.04  &6.49   &0.97  \\
		MDP-Flow2~\cite{xu2012motion} 				&2.89 	 & {0.59}  &{3.47} 	 &{0.62}   	&3.66 		 &1.24  	&{5.20} 	  &	{0.94} 	&10.2 & 	0.98 &6.13 	 &	1.09  	&7.36 & 0.70  	&7.75 & 	0.78 	&5.83   &	0.87 	 \\
		DeepFlow2~\cite{weinzaepfel2013deepflow}	&2.98 	 &0.62  &3.88 	 &0.74   	&{3.62} 		 &{0.86}  	&5.39 	  &	0.99 	&11.0 & 	1.04 &5.91 	 &	1.02  	&7.14 & 0.63  	&7.80 & 	0.96 	&5.97   &	0.86 	 \\
		SepConv-$L_1$~\cite{niklaus2017videoSepConv} &{2.52} 	 &{0.54}  &3.56 	 &0.67   	&4.17 		 &1.07  	&5.41 	  &	1.03 	&10.2 & 	0.99 &5.47 	 &	{0.96}  	&6.88 & 0.68  	&{6.63} & 	0.70 	&5.61   &	0.83 	 \\
		
		SuperSlomo~\cite{jiang2017super}	&{2.51} 	 &0.59  &3.66 	 &0.72   	&\first{{2.91}} 		 &\first{{0.74}}  	&{5.05} 	  &	0.98 	&9.56  &{0.94} &5.37 	 &	{0.96}  	&6.69 & \first{0.60}  	&6.73 & 	{0.69} 	&{5.31}   &	{0.78} 	 \\
		
		CtxSyn~\cite{niklaus2018context}	&\first{{2.24}} 	 &\first{{0.50}}  &\first{{2.96}} 	 &\first{{0.55}}   	&{4.32} 		 &{1.42}  	&\first{{4.21}} 	  &	\first{{0.87}}	&9.59  &	0.95 &5.22 	 &	{0.94}  	&7.02 & 0.68  	&6.66 & 	{0.67} 	&{5.28}   &	{0.82} 	 \\

		\Ours	&2.83 	 &0.64  &3.84 	 &0.73   	&4.16 		 &{0.84}  	&5.75 	  &	0.99 	&{{8.57}}  &	{{0.93}} &{{4.99}} 	 &{{0.96}}  	&\first{{5.86}} & \first{{0.60}} 	&{6.83} & {{0.69}} 	&{5.35}   &	{0.80} 	 \\
		

		\Ours* &2.39 &	0.59	&3.36  &0.64 	&3.37 &0.80  	&4.84 &0.88 	&\first{8.55} 	& \first{0.88}  &\first{4.70} & \first{0.85}  	&6.40 &0.64 	&\first{6.37} 	  &	\first{0.63} &\first{5.00} 	& \first{0.74}  \\

		\bottomrule
	\end{tabular} 
	
\end{table*}

\subsection{Video Frame Interpolation}
We first provide the comparison with the sequential model and then present quantitative and qualitative evaluations with the state-of-the-art approaches.
\begin{table}[!t]
	\caption{\textbf{Average IE and NIE values with standard variances on Middlebury benchmark.}}
	\vspace{-8pt}
	\label{tab:mean_std_of_IE_NIE}
	\footnotesize
	\centering
	\begin{tabular}{ccc}
		\toprule
		Method          & IE & NIE \\ 
		\midrule
		EpicFlow~\cite{revaud2015epicflow}  & 6.48 $\pm$ 2.75 & 0.97 $\pm$ 0.20 \\
		MDP\_Flow2~\cite{xu2012motion}      & 5.83 $\pm$ 2.52 & 0.86 $\pm$ 0.23 \\
		DeepFlow2~\cite{weinzaepfel2013deepflow}    & 5.96 $\pm$ 2.65 & 0.85 $\pm$ 0.17 \\
		SepConv\_L$_1$~\cite{niklaus2017videoSepConv}  & 5.60 $\pm$ 2.37 & 0.83 $\pm$ 0.20 \\
		SuperSlomo~\cite{jiang2017super}      & 5.31 $\pm$ 2.34 & \second{0.77 $\pm$ 0.16} \\
		CtxSyn~\cite{niklaus2018context}          & \second{5.27 $\pm$ 2.39} & 0.82 $\pm$ 0.29 \\
		MEMC-Net        & 5.35 $\pm$ 1.81 & 0.79 $\pm$ 0.15 \\
		MEMC-Net*       & \first{4.99 $\pm$ 2.02} & \first{0.73 $\pm$ 0.12} \\
		\bottomrule
	\end{tabular} 
\end{table}

\begin{figure*}[t]
	\footnotesize
	\centering
\renewcommand{\tabcolsep}{0.1pt} 
\renewcommand{\arraystretch}{0.1} 
\begin{center}
	\begin{tabular}{ccccccc}
 \multirow{1}{*}[9.6em]{	
\includegraphics[width=0.20\linewidth]{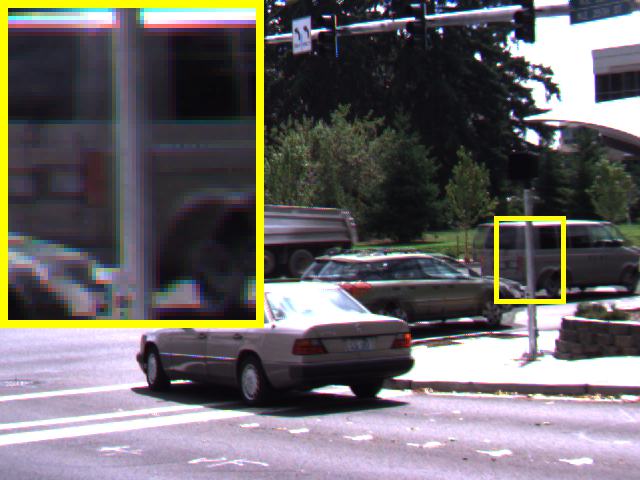}
}
&
\includegraphics[width=0.125\linewidth]{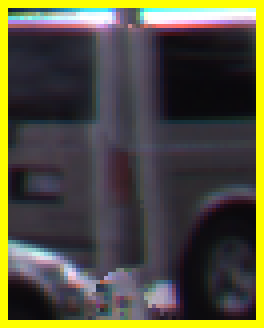}&
\includegraphics[width=0.125\linewidth]{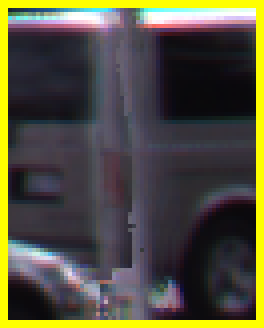}&
\includegraphics[width=0.125\linewidth]{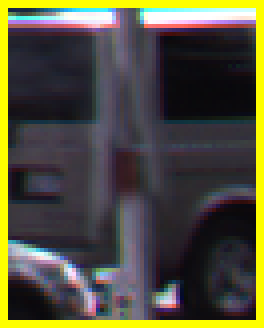}&
\includegraphics[width=0.125\linewidth]{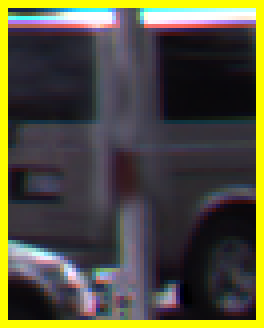}&
\includegraphics[width=0.125\linewidth]{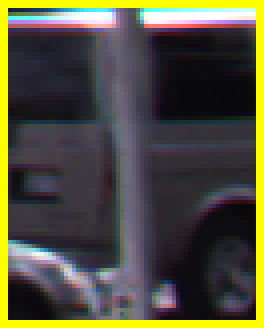} &
\includegraphics[width=0.125\linewidth]{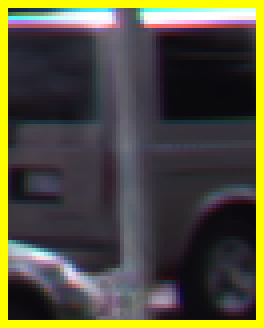} \\

\addlinespace[0.2cm]

 \multirow{2}{*}[1.45em]{	
\includegraphics[width=0.20\linewidth]{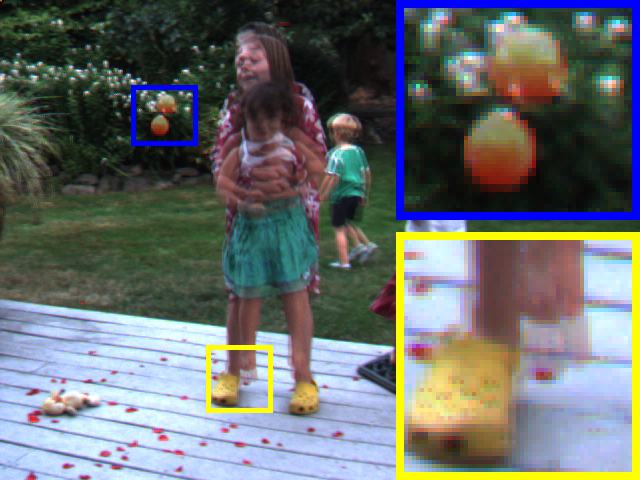}
}
&
\includegraphics[width=0.125\linewidth]{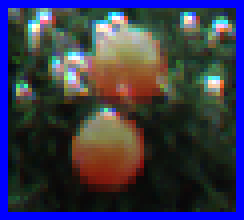} &
\includegraphics[width=0.125\linewidth]{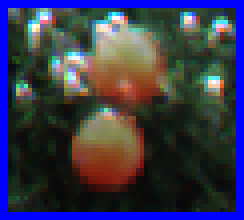}&
\includegraphics[width=0.125\linewidth]{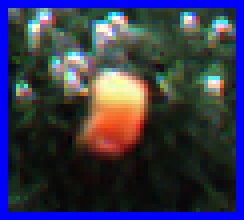}&
\includegraphics[width=0.125\linewidth]{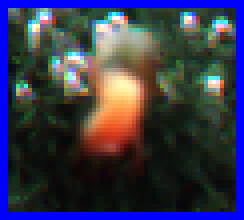}&
\includegraphics[width=0.125\linewidth]{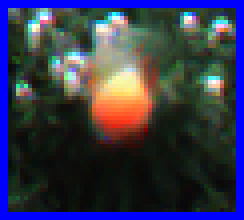}&
\includegraphics[width=0.125\linewidth]{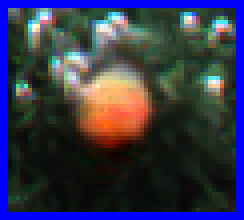}
\\	
&
\includegraphics[width=0.125\linewidth]{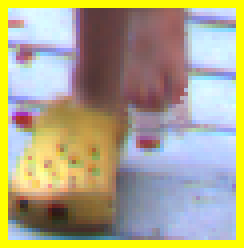} &
\includegraphics[width=0.125\linewidth]{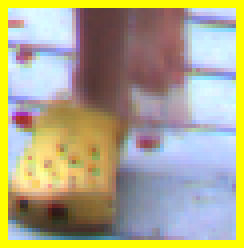}&
\includegraphics[width=0.125\linewidth]{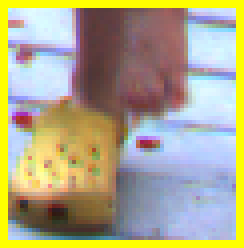}&
\includegraphics[width=0.125\linewidth]{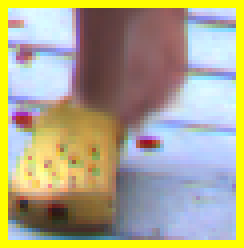}&
\includegraphics[width=0.125\linewidth]{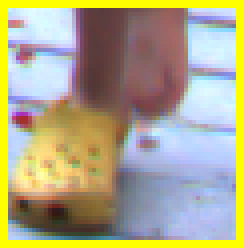}&
\includegraphics[width=0.125\linewidth]{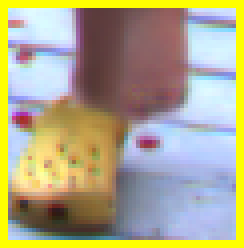}
\\

		\addlinespace[0.2cm]
			(a) Overlay &
			(b) EpicFlow&
			(c) SPyNet&
			(d) SepConv-$L_f$& 
			(e) SepConv-$L_1$&
			(f) \Ours &
			(g) \Ours* \\

		\end{tabular}
	\end{center}
	\vspace{-0.3cm}
	\caption{
	\textbf{Visual comparisons on Middlebury~\cite{baker2011database}.}
	The sequences are from the \textsc{Evaluation} set.
	}
\label{fig:Middlebury} 
\end{figure*}

\begin{figure}[!t]

	\centering	
	\begin{minipage}{0.49\linewidth}
		\centering{	 \includegraphics[width= 1.0\textwidth]{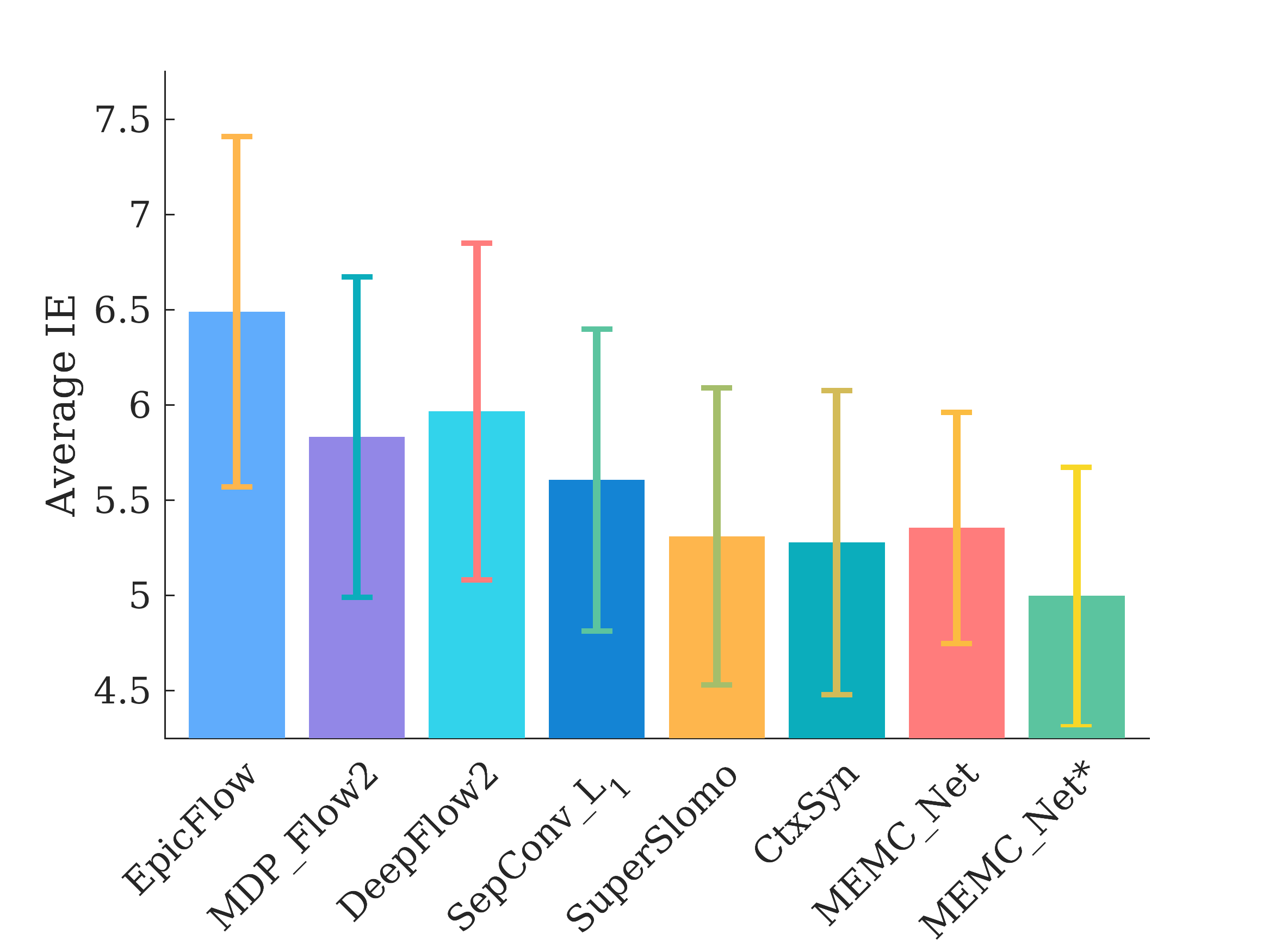}}
	\end{minipage}
		\begin{minipage}{0.49\linewidth}
			\centering{	 \includegraphics[width= 1.0\textwidth]{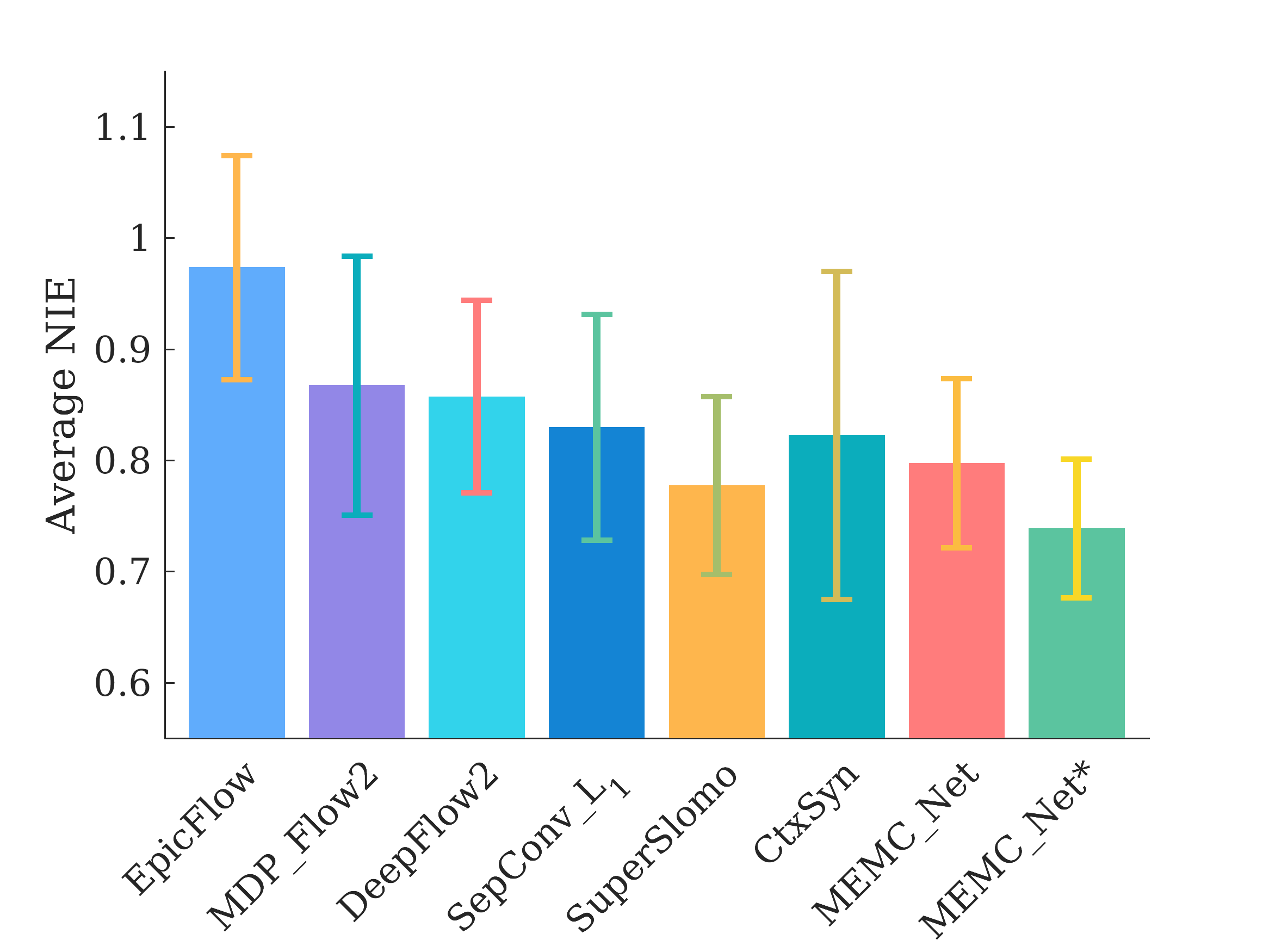}}			
		\end{minipage}
		
	\vspace{-5pt}
	\caption{\textbf{Error bars on the Middlebury sequences.} 
	 }
	\label{fig:errorbar}
\end{figure}

\begin{table*}[!t]
 \caption{
 	\textbf{Quantitative evaluation on UCF101, Vimeo90K, and Middlebury datasets.}
 	The abbreviations \textit{oth.} and \textit{eval.} represent the \textsc{Other} and \textsc{Evaluation} sets in the Middlebury dataset.
 	The numbers in \first{red} depict the best performance, while the numbers in \second{blue} depict the second-best performance.
 }
 \label{tab:UCF101}
\footnotesize
\newcommand{\tabincell}[2]{\begin{tabular}{@{}#1@{}}#2\end{tabular}}
\newcolumntype{P}[1]{>{\centering\arraybackslash}p{#1}}

\centering
\begin{tabular}{m{2.5cm}  P{1.1cm} P{1.1cm}  P{1.1cm} P{1.1cm}  P{1.4cm} P{1.4cm} }
\toprule
\multirow{2}{*}[-0.2em]{Methods} &  \multicolumn{2}{c}{UCF101~\cite{soomro2012ucf101}} &\multicolumn{2}{c}{Vimeo90K~\cite{xue2017video}} &\multicolumn{2}{c}{Middlebury~\cite{baker2011database}}  \\
\cmidrule(l{2pt}r{2pt}){2-3}
\cmidrule(l{2pt}r{2pt}){4-5}
\cmidrule(l{2pt}r{2pt}){6-7}
	&PSNR & SSIM 	&PSNR & SSIM	& 
	IE \scriptsize{(\textit{oth.})} & IE \scriptsize{(\textit{eval.})} \\
\midrule
SPyNet~\cite{ranjan2017optical} & 33.67 & 0.9633 &31.95 & 0.9601& 2.49& --- \\
EpicFlow~\cite{revaud2015epicflow} &33.71 & 0.9635 &32.02&0.9622& 2.47& 6.48  \\
MIND~\cite{long2016learning} & 33.93 & 0.9661 & 33.50& 0.9429 & 3.35&--- \\
DVF~\cite{liu2017video} & 34.12 & 0.9631 & 31.54 &  0.9462 & 7.75&---  \\
ToFlow~\cite{xue2017video} & 34.54 & 0.9666 &33.53 &0.9668& --- & ---  \\
ToFlow+Mask~\cite{xue2017video} & 34.58 & 0.9667  &33.73& 0.9682 &2.51& --- \\
SepConv-$L_f$~\cite{niklaus2017videoSepConv} & 34.69 & 0.9655  & 33.45& 0.9674& 2.44& ---  \\
SepConv-$L_1$~\cite{niklaus2017videoSepConv} & 34.78 & 0.9669  &33.79& {0.9702}& {2.27} & 5.61  \\

\Ours\_s & 34.83 & {0.9676} &{33.97} &{0.9721} &{2.43} & --- \\ 

\Ours & \second{34.95} & \second{0.9679} &\second{34.02}& \second{0.9704} &\second{2.24} &  \second{5.35}  \\

\Ours* & \first{35.01} & \first{0.9683} &\first{34.40}& \first{0.9742} &\first{2.10} &  \first{5.00}  \\

\bottomrule
\end{tabular} 

\end{table*}

\begin{table*}[!t]
 
 \caption{\textbf{Quantitative evaluation on HD videos.}}
 \label{tab:HDdataset}
 
\footnotesize
\newcommand{\tabincell}[2]{\begin{tabular}{@{}#1@{}}#2\end{tabular}}
\newcolumntype{P}[1]{>{\centering\arraybackslash}p{#1}}

\centering
\begin{tabular}{m{1.3cm}  P{1.4cm}  P{1.0cm} P{1.0cm}  P{1.0cm} P{1.0cm}  P{1.0cm} P{1.0cm} P{1.0cm} P{1.0cm} P{1.0cm} P{1.0cm}}
\toprule
\multirow{2}{*}[-0.2em]{Video} & 
\multirow{2}{*}[-0.2em]{Resolution} &
\multicolumn{2}{c}{ToFlow+Mask~\cite{xue2017video}} & 
\multicolumn{2}{c}{SepConv-$L_f$~\cite{niklaus2017videoSepConv} } &
\multicolumn{2}{c}{SepConv-$L_1$~\cite{niklaus2017videoSepConv}} &
\multicolumn{2}{c}{\Ours}   &
\multicolumn{2}{c}{\Ours*}  
\\
\cmidrule(l{2pt}r{2pt}){3-4} 
\cmidrule(l{2pt}r{2pt}){5-6}
\cmidrule(l{2pt}r{2pt}){7-8}
\cmidrule(l{2pt}r{2pt}){9-10}
\cmidrule(l{2pt}r{2pt}){11-12}
&	&PSNR & SSIM 	&PSNR & SSIM	& PSNR & SSIM	& PSNR & SSIM& PSNR & SSIM	\\
\midrule

Alley2 &  544p    	 &26.30	&0.7997&	28.26&	0.8462&	{28.52}&{0.8646}&	 \second{29.57} &	\second{0.8845} &\first{29.60} & 	\first{0.8920} \\
Market5 & 544p   	 &18.21	&0.7324&	20.59&	0.7878&	20.57&	{0.8012}&	 \second{21.16}&	\second{0.8074} &\first{21.64}& \first{0.8105}\\
Temple1 & 544p   	 &25.20	&0.9174&	26.42&	0.9295&	26.69&	\second{0.9370}&	 \first{27.25}&	0.9354   &{27.18}&  \first{0.9390}\\
Temple2 & 544p   	 &19.90	&0.8246&	21.74&	0.8471&	21.93&	\second{0.8533}&	 \second{22.72}&	\first{0.8628} &\first{22.94}& {0.8532}\\

Parkrun & 720p 	     &27.77	&0.8841&    28.69&	0.9083&	{29.03}&	\first{0.9158}&	 \second{29.07}&	0.9125 &\first{29.15}& \second{0.9145}\\
Shields & 720p	 	 &34.10	&0.8884&	34.55&	0.9093&	34.91&	 {0.9188}&	  \second{35.21}&	\second{0.9206} &\first{35.49}& \first{0.9251}\\
Stockholm & 720p 	 &33.53	&0.8534&	33.99&	0.8669&	34.27&	0.8826&	 \second{34.64}&\second{0.8894} &\first{34.89}&	 \first{0.8931}\\

Kimono & 1080p   	 &33.34	&0.9107&	34.07&	0.9168&	34.31&	0.9287&	 \second{34.93}&	\second{0.9341} &	 \first{34.99}&	\first{0.9363}\\
ParkScene & 1080p	 &33.49	&0.9233&	35.27&	0.9374&	35.51&	{0.9451}& \second{36.20}&\second{0.9491} & \first{36.64}& \first{0.9521}\\
Sunflower & 1080p	 &33.75	&0.9476&	34.88&	0.9539&	35.02&	0.9605&	 \second{35.42}&	\second{0.9616}&	 \first{35.59}&	\first{0.9638} \\
Bluesky & 1080p  	 &37.53	&0.9673&	38.32&	0.9730&	38.83&	0.9775&	 \second{39.28}&	\second{0.9791} &	 \first{39.55}&	\first{0.9801} \\
\midrule
Average && 29.37& 0.8772 &30.61 & 0.8978 &30.87 &{0.9077} & \second{31.40} &\second{0.9124} &\first{31.60} &\first{0.9145}\\

\bottomrule
\end{tabular} 

\end{table*}

\begin{figure*}[t]
	\footnotesize
	\centering
	\renewcommand{\tabcolsep}{2pt} 
	\renewcommand{\arraystretch}{1} 
	\begin{center}
		\begin{tabular}{ccccc}
\includegraphics[width=0.19\linewidth]{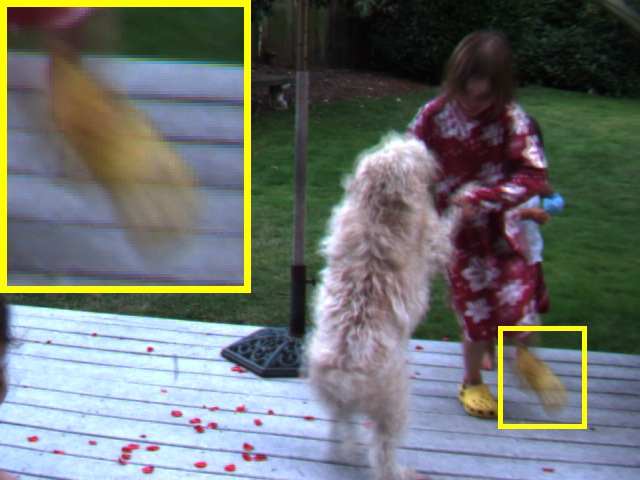}&
\includegraphics[width=0.19\linewidth]{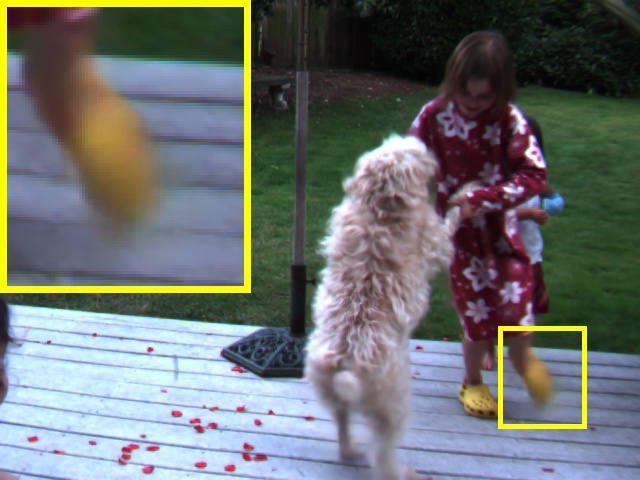}&
\includegraphics[width=0.19\linewidth]{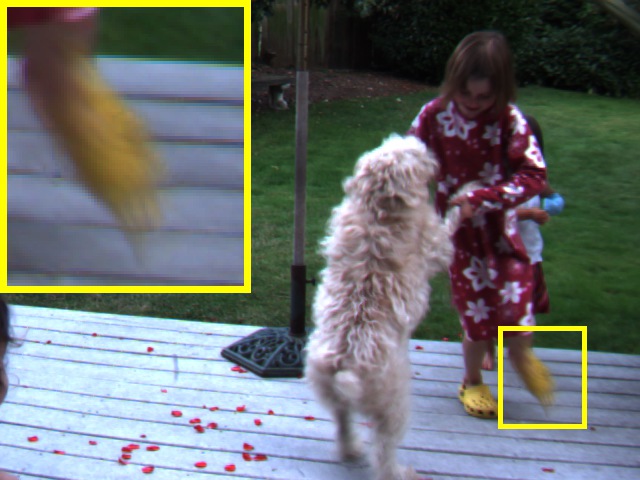}&
\includegraphics[width=0.19\linewidth]{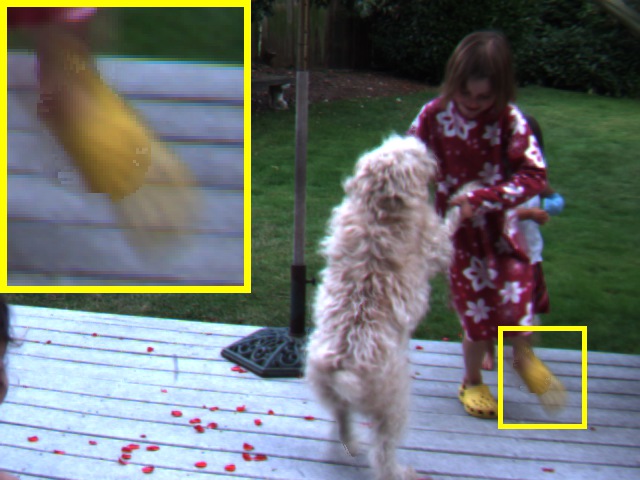}&
\includegraphics[width=0.19\linewidth]{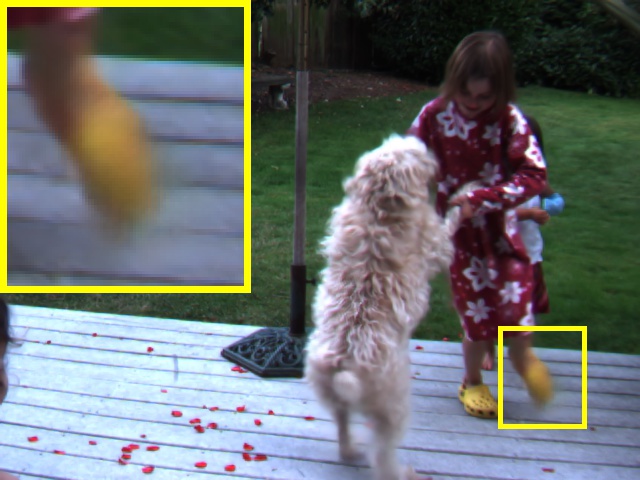}
 \\
(a) Overlay &
(b) MIND~\cite{long2016learning}&
(c) ToFlow~\cite{xue2017video}&
(d) EpicFlow~\cite{revaud2015epicflow}&
(e) SPyNet~\cite{ranjan2017optical} 
\\

\includegraphics[width=0.19\linewidth]{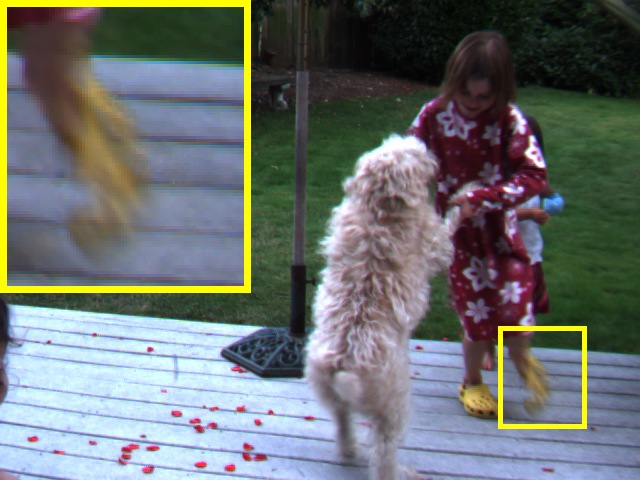} &
\includegraphics[width=0.19\linewidth]{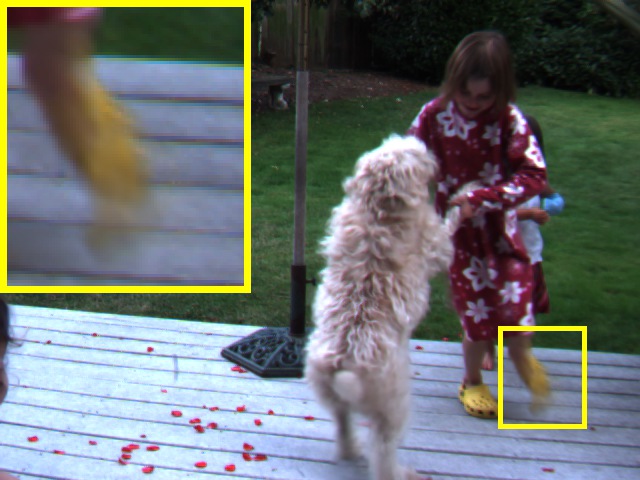} &
\includegraphics[width=0.19\linewidth]{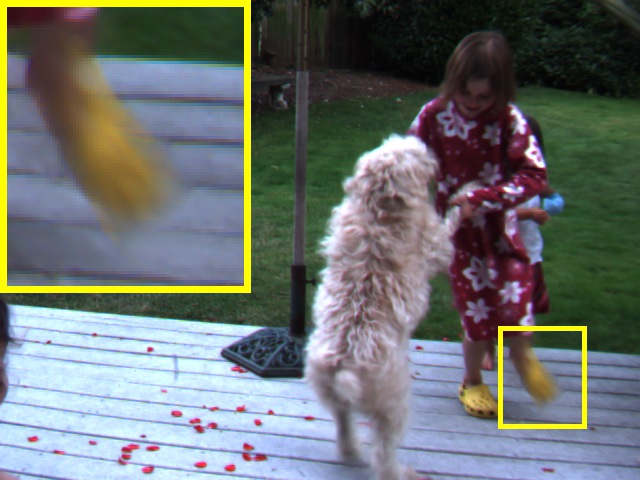} &
\includegraphics[width=0.19\linewidth]{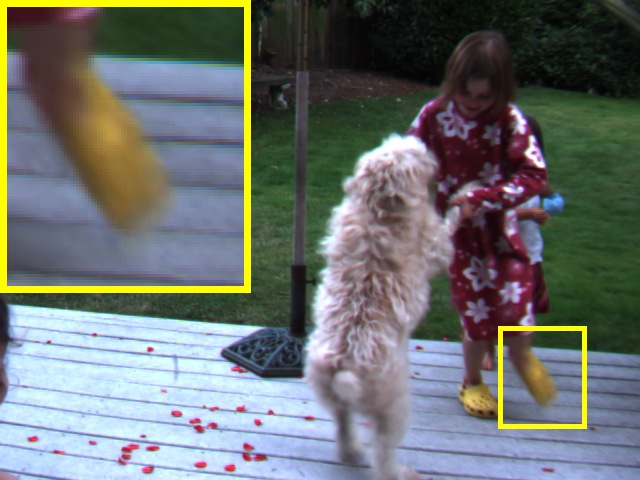} &
\includegraphics[width=0.19\linewidth]{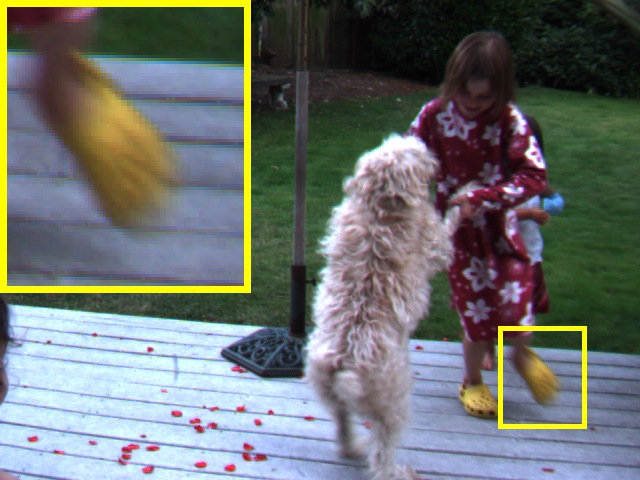}
\\
			 
(f) SepConv-$L_f$~\cite{niklaus2017videoSepConv}&
(g) SepConv-$L_1$~\cite{niklaus2017videoSepConv}&
(h) \Ours~ &
(i) \Ours*~ &
(j) Ground Truth\\	
		\end{tabular}
	\end{center}
	\vspace{-0.4cm}
	\caption{
	\textbf{Visual comparisons on Middlebury~\cite{baker2011database}.}
	The sequences are from the \textsc{Other} set.
	}
\label{fig:Middlebury-other} 
\end{figure*}

\begin{figure*}[t]
		\footnotesize
	\centering
	\renewcommand{\tabcolsep}{1pt} 
	\renewcommand{\arraystretch}{1} 
\begin{center}
	\begin{tabular}{cccccc}
\includegraphics[width=0.161\linewidth]{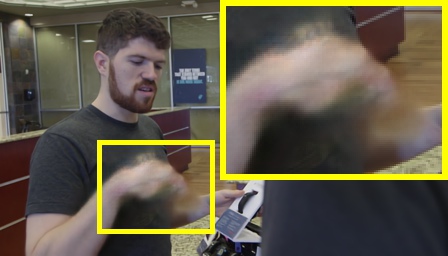}&
\includegraphics[width=0.161\linewidth]{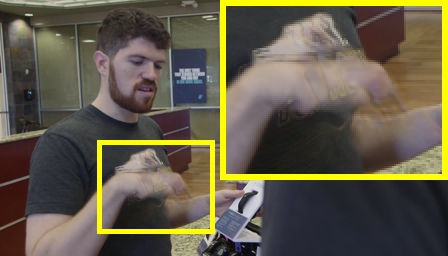}&
\includegraphics[width=0.161\linewidth]{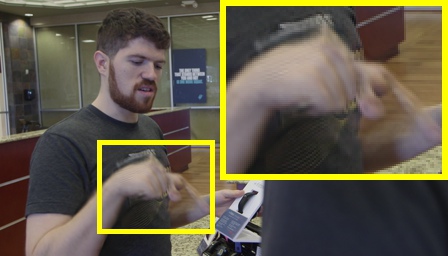}&
\includegraphics[width=0.161\linewidth]{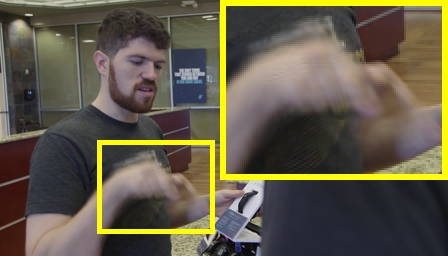}&
\includegraphics[width=0.161\linewidth]{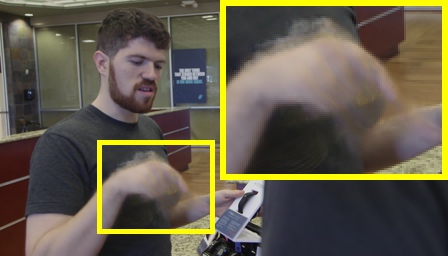}&
\includegraphics[width=0.161\linewidth]{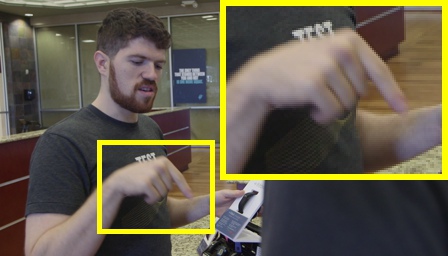} \\

\includegraphics[width=0.161\linewidth]{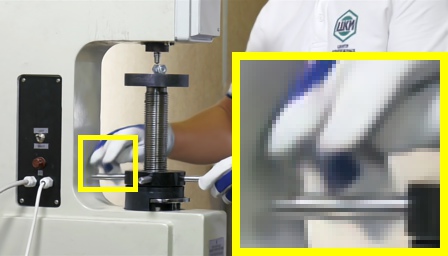}&
\includegraphics[width=0.161\linewidth]{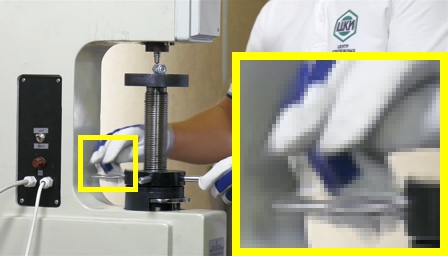}&
\includegraphics[width=0.161\linewidth]{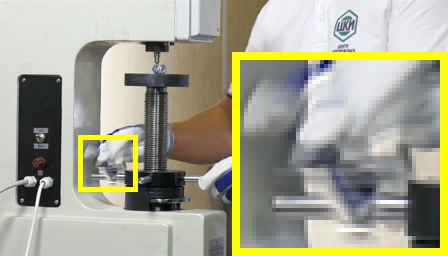}&
\includegraphics[width=0.161\linewidth]{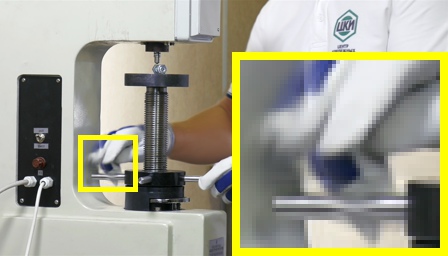}&
\includegraphics[width=0.161\linewidth]{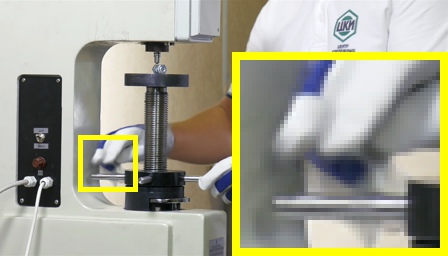}&
\includegraphics[width=0.161\linewidth]{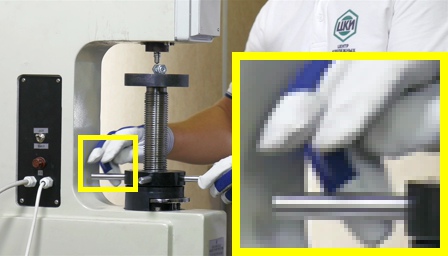} \\	

\includegraphics[width=0.161\linewidth]{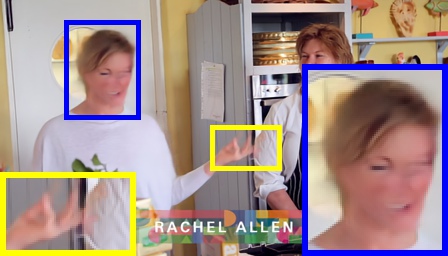}&
\includegraphics[width=0.161\linewidth]{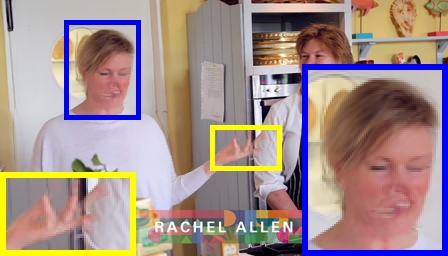}&
\includegraphics[width=0.161\linewidth]{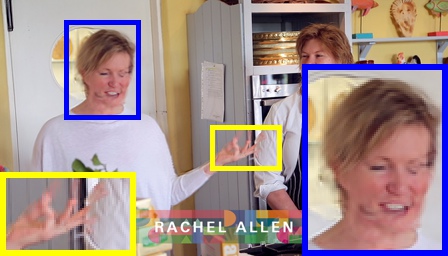}&
\includegraphics[width=0.161\linewidth]{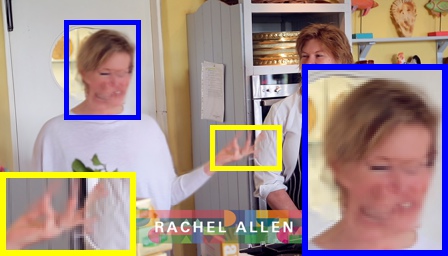}&
\includegraphics[width=0.161\linewidth]{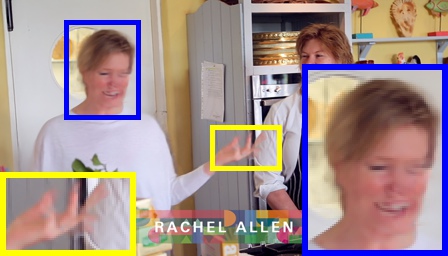}&
\includegraphics[width=0.161\linewidth]{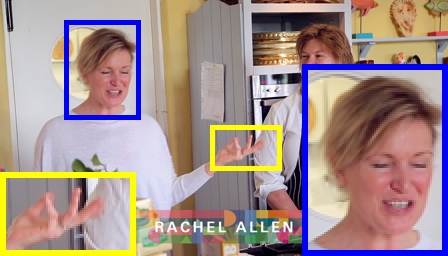} \\	

(a) MIND~\cite{long2016learning}&
(b) ToFlow~\cite{xue2017video}&
(c) SepConv-$L_f$\cite{niklaus2017videoSepConv}& 
(d) SepConv-$L_1$\cite{niklaus2017videoSepConv}&
(e) \Ours&
(f) Ground Truth\\	
		\end{tabular}
	\end{center}
	\vspace{-0.4cm}
\caption{\textbf{Visual comparisons on the Vimeo90K dataset~\cite{xue2017video}.}}
\label{fig:Vimeo} 
\vspace{-12pt}
\end{figure*}

\subsubsection{Comparisons with the Sequential Model}
\label{sec:SM}
{We train the sequential model and present the quantitative results in~\tabref{sequentialmodel}.
Compared to the proposed approach, the performance of the sequential model is 0.61dB and 1.08dB lower on the UCF101 and Vimeo90K datasets, respectively.
%
We attribute the performance difference to the flow warping errors on the motion boundary (which has occlusion and dis-occlusion). 
The proposed model avoid estimating kernels from warped images and leads to better performance.}

\subsubsection{Comparisons with the State-of-the-arts}
We evaluate the proposed \Ours\ and \Ours* against the kernel-based method (SepConv~\cite{niklaus2017videoSepConv}), flow-based algorithms (DVF~\cite{liu2017video}, ToFlow~\cite{xue2017video}, and CtxSyn~\cite{niklaus2018context}), and a direct interpolation approach (MIND~\cite{long2016learning}).
The ToFlow method~\cite{xue2017video} generates two results with and without learning occlusion masks.
Two pre-trained models of the SepConv approach~\cite{niklaus2017videoSepConv} are
available: the SepConv-$L_1$ model is optimized with a $L_1$ loss function while the SepConv-$L_f$ model uses both the $L_1$ loss and the perceptual loss~\cite{simonyan2014very} for generating more realistic results.
As no pre-trained model of the MIND method~\cite{long2016learning} is available, 
we train their network on the Vimeo90K training set for evaluations.
In addition to the above learning-based frame interpolation methods, we also use existing optical flow algorithms (SPyNet~\cite{ranjan2017optical} and EpicFlow~\cite{revaud2015epicflow}) to directly interpolate frames.

We show the interpolation results of the Middlebury \textsc{Evaluation} set in~\tabref{Middlebury}.
%
%
These results are also publicly available on the Middlebury benchmark website (\url{http://vision.middlebury.edu/flow/eval/results/results-i1.php}).
%
{For the sequences with smaller motions (i.e., the maximum flow magnitude is less than 10 pixels) or fine textures, such as the \textit{Mequon}, \textit{Teddy} and \textit{Schefflera}, the CtxSyn method~\cite{niklaus2018context} obtains the best results in terms of both  IE and NIE metrics.
  			In contrast, our MEMC-Net and MEMC-Net* perform well on the videos with complicated motion, e.g., the \textit{Backyard} sequence with dancing feet and the \textit{Basketball} video with moving arms and fingers.
  			Notably, the SuperSlomo method~\cite{jiang2017super} generates the best results for the synthetic \textit{Urban} sequence.
  			%
  			%
  			On average, the proposed models perform favorably against the state-of-the-art approaches on the Middlebury dataset.
}
\modify{In~\tabref{mean_std_of_IE_NIE}, we present the average IE and NIE values with standard variances on the Middlebury dataset.}
{Also in~\figref{errorbar}, we present error bars on the IE and NIE metrics to show the statistical comparison between different methods.
The MEMC-Net* obtains lower interpolation error at smaller variance on different scenarios}.

\begin{figure*}[!ht]
	\footnotesize
	\centering
	\renewcommand{\tabcolsep}{1pt} 
	\renewcommand{\arraystretch}{1} 
	\begin{center}
		\begin{tabular}{cccccc}
			
			\includegraphics[width=0.16\linewidth]{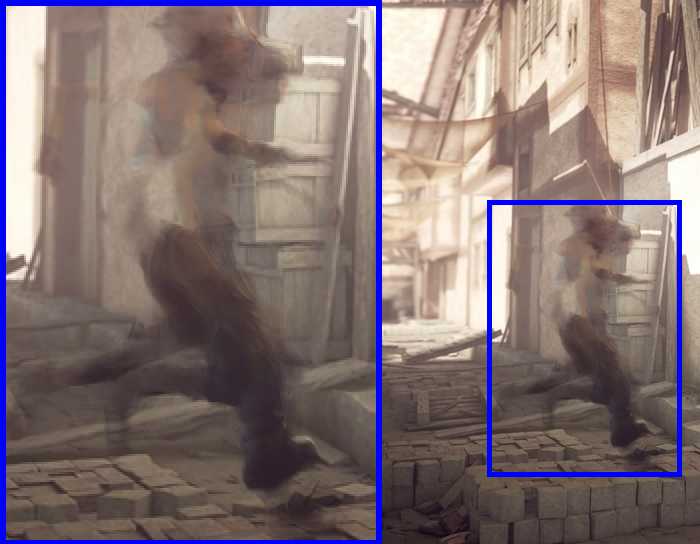}&
			\includegraphics[width=0.16\linewidth]{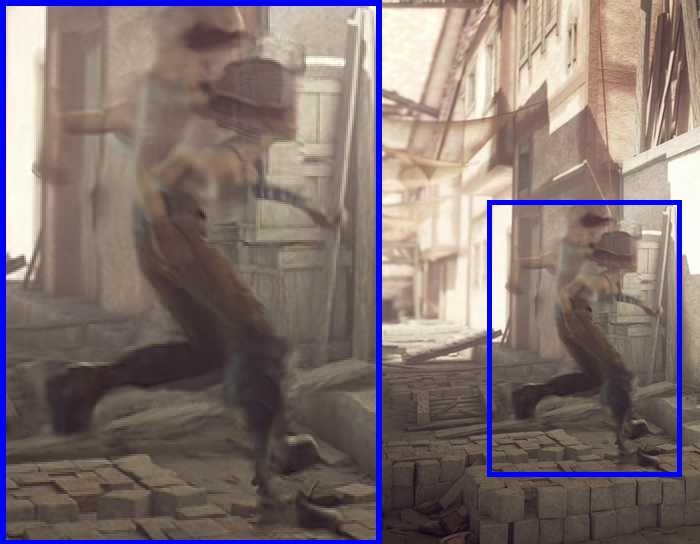}&
			\includegraphics[width=0.16\linewidth]{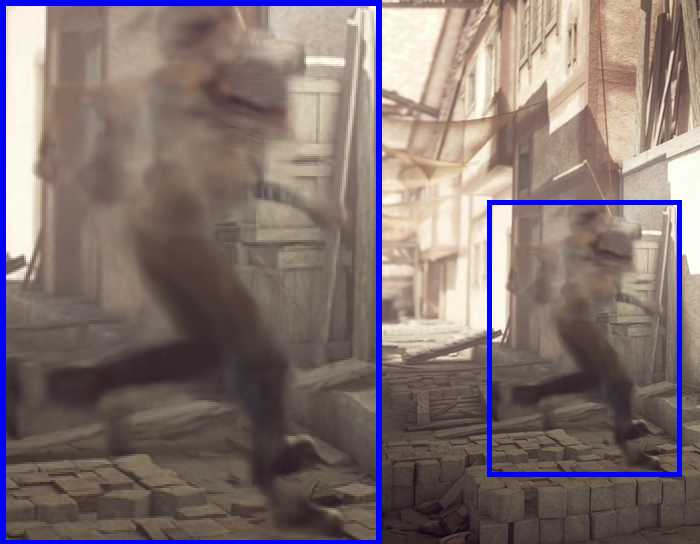}&		
			\includegraphics[width=0.16\linewidth]{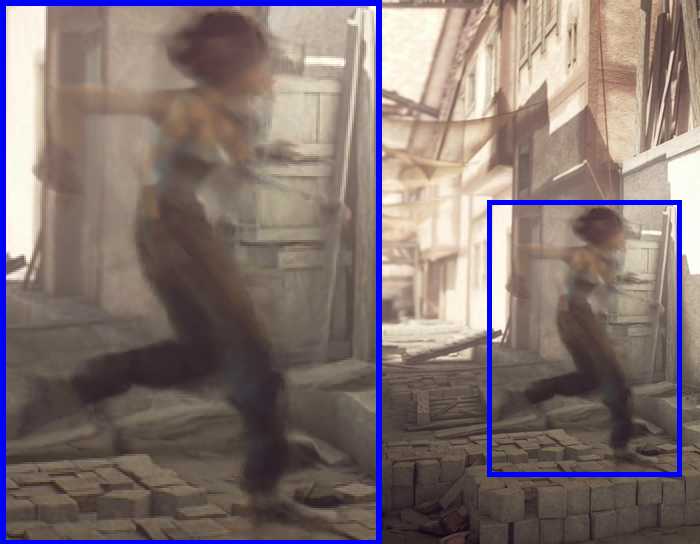} &
			\includegraphics[width=0.16\linewidth]{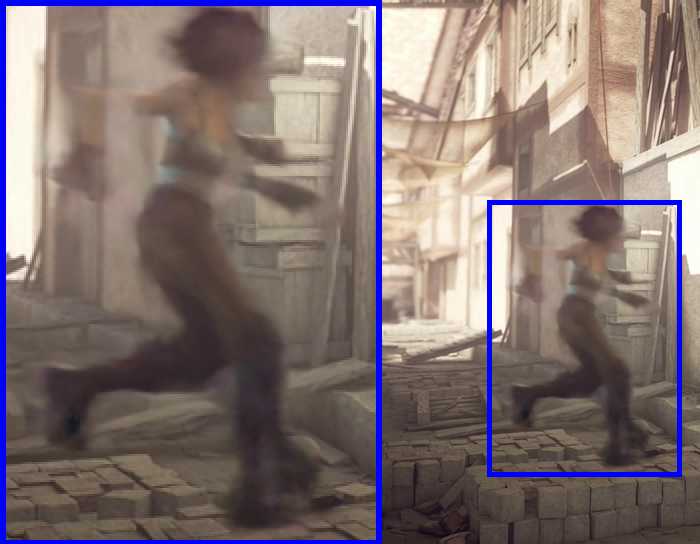} &
		    \includegraphics[width=0.16\linewidth]{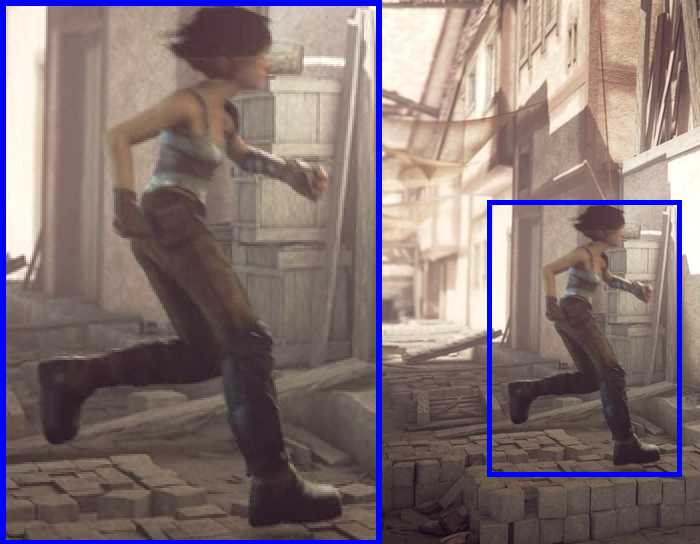}\\
			(a) ToFlow~\cite{xue2017video}&
			(b) SepConv-$L_f$~\cite{niklaus2017videoSepConv}&
		    (c) SepConv-$L_1$~\cite{niklaus2017videoSepConv}&
			(d) \Ours &
			(e) \Ours* &
			(f) Ground Truth \\
		\end{tabular}
	\end{center}
	\vspace{-0.4cm}
	\caption{
		\textbf{Visual comparisons on HD videos.}
		%
	}
	\label{fig:HD-Alley2} 
	\vspace{-0.2cm}
\end{figure*}

\tabref{UCF101} shows that the proposed methods perform favorably against the state-of-the-art approaches on the UCF101~\cite{soomro2012ucf101}, Vimeo90K~\cite{xue2017video}, and Middlebury~\cite{baker2011database} datasets.
The numbers in \first{red} depict the best performance, {while the numbers in \second{blue} depict the second-best performance}.
The diverse scenarios in these video datasets demonstrate that our model generalizes well to different types of motion.
On the other hand, the MIND model~\cite{long2016learning} trained on the same Vimeo90k training set does not perform well on the UCF101 and Middlebury datasets.

In~\tabref{HDdataset}, we present the evaluation results on the HD videos, which typically contain much {larger motion}.
Our approach consistently performs well against the state-of-the-art methods on different resolutions.
The performance gap between the proposed \Ours{} and SepConv~\cite{niklaus2017videoSepConv} becomes larger especially on 1080p videos, which demonstrates that it is not feasible 
to handle large motion with the fixed kernel size (e.g., $51 \times 51$).
Our \Ours* with context information and residual blocks performs favorably against
 the existing methods with significant improvement up to 0.9dB (e.g., \textit{Alley2} and \textit{ParkScene}).

\subsubsection{Qualitative Results}
We present sample interpolation results from the evaluated datasets in~\figref{Middlebury}, \ref{fig:Middlebury-other}, \ref{fig:Vimeo}, and \ref{fig:HD-Alley2}.
%
On the first row of~\figref{Middlebury}, the EpicFlow~\cite{revaud2015epicflow} and SPyNet~\cite{ranjan2017optical} methods do not reconstruct the straight lamppost 
due to inaccurate optical flow estimation.
Both the SepConv-$L_1$ and SepConv-$L_f$~\cite{niklaus2017videoSepConv} models cannot interpolate the lamppost well as the motion is larger than the size of the interpolation kernels.
In contrast, our method reconstructs the lamppost well.
On the second row of~\figref{Middlebury}, the proposed method interpolates the falling ball with a clear shape with fewer artifacts on the leg of the girl.

In \figref{Vimeo}, the ToFlow~\cite{xue2017video} and SepConv~\cite{niklaus2017videoSepConv} methods generate ghost effect around the hand.
Due to large and non-rigid motion, 
flow-based methods are less effective in estimating accurate optical flow for interpolation, 
while kernel-based approaches are not able to infer motion beyond the size of local kernels.
In contrast, our model reconstructs the hand with fewer visual artifacts.
As shown in \figref{HD-Alley2}, the SepConv~\cite{niklaus2017videoSepConv} method is not able 
to interpolate the scene well due to large motion. 
In contrast, the proposed method interpolates frames well with visually pleasing results when 
optical flow is not accurately estimated.
\begin{table}[!t]
\begin{minipage}{\linewidth}
 \caption{
 \textbf{Quantitative evaluation for video super-resolution.}
 }
 \vspace{-8pt}
 \label{tab:super-resolution}
 \footnotesize
\newcommand{\tabincell}[2]{\begin{tabular}{@{}#1@{}}#2\end{tabular}}
\newcolumntype{P}[1]{>{\centering\arraybackslash}p{#1}}

\begin{tabular}{c  m{2.0cm}  P{0.8cm} P{0.8cm}  P{0.8cm} P{0.8cm}}
\toprule
\multirow{3}{*}{\tabincell{c}{Frame \\ \#Num.}}& \multirow{3}{*}[0.1em]{Methods} &\multicolumn{2}{c}{Vimeo90K~\cite{xue2017video}}  & \multicolumn{2}{c}{BayesSR~\cite{liu2011bayesian}} \\
\cmidrule(l{2pt}r{2pt}){3-4} \cmidrule(l{2pt}r{2pt}){5-6}
	&&PSNR & SSIM	&PSNR & SSIM 	 \\
	\midrule

\multirow{2}{*}{1} &Bicubic &29.79 &0.9036& 22.17 &0.7391\\
&EDSR~\cite{lim2017enhanced}& \second{33.08} &0.9411 &\second{23.93}&\second{0.8113}\\
\midrule
\multirow{5}{*}[-0.5em]{7}
&DeepSR~\cite{liao2015video}& 25.55 &0.8498 &21.85 & 0.7535\\
&BayesSR~\cite{liu2011bayesian}& 24.64 &0.8205 & 21.95 & 0.7369  \\
&ToFlow~\cite{xue2017video} &\second{33.08} & \second{0.9417} & {23.54} & {0.8070} \\
\cmidrule{2-6}

&\Ours{\_SR} & \first{33.47} &\first{0.9470}   & \first{24.37}& \first{0.8380}\\


\bottomrule
\end{tabular} 
\end{minipage}	 

\vfill
\vspace{0.4cm}

\begin{minipage}{\linewidth}
\caption{ \textbf{Quantitative evaluation for video denoising.}}
 \vspace{-8pt}
\label{tab:denoising}
\footnotesize
\newcommand{\tabincell}[2]{\begin{tabular}{@{}#1@{}}#2\end{tabular}}
\newcolumntype{P}[1]{>{\centering\arraybackslash}p{#1}}

\centering
\begin{tabular}{c  m{2.1cm}  P{0.8cm} P{0.8cm}  P{0.8cm} P{0.8cm}}
	\toprule
\multirow{3}{*}{\tabincell{c}{Frame \\ \#Num.}}&
\multirow{2}{*}[-0.3em]{Methods} &\multicolumn{2}{c}{Vimeo90K~\cite{xue2017video}}  
&\multicolumn{2}{c}{V-BM4D~\cite{maggioni2012video}} \\
\cmidrule{3-6}
&	&PSNR & SSIM 	&PSNR & SSIM \\
	\midrule
\multirow{2}{*}{1}
&Noisy & 22.63 & 0.5007 & 22.28 & 0.4715 \\

&EDSR\_DN~\cite{lim2017enhanced}& \second{35.11}  &\second{0.9513}  & 32.02 & 0.8828  \\

\midrule
\multirow{3}{*}[-1em]{7}
&	ToFlow~\cite{xue2017video} & 32.66 &  0.9198 & 30.19 &  0.8699  \\

& V-BM4D~\cite{maggioni2012video} &34.39 &0.9217&\second{32.27}&\second{0.8913}  \\

 	\cmidrule{2-6}

&	\Ours{\_DN}  & \first{36.35} & \first{0.9642} &\first{34.22}& \first{0.9310}\\
	
	\bottomrule
\end{tabular} 
\end{minipage}

\vfill
\vspace{0.4cm}

\begin{minipage}{\linewidth}
\caption{ \textbf{Quantitative evaluation for video deblocking.} }
 \vspace{-8pt}
\label{tab:deblocking}
\footnotesize
\newcommand{\tabincell}[2]{\begin{tabular}{@{}#1@{}}#2\end{tabular}}
\newcolumntype{P}[1]{>{\centering\arraybackslash}p{#1}}

\centering
\begin{tabular}{c  m{2.0cm}  P{0.8cm} P{0.8cm}  P{0.8cm} P{0.8cm}}
	\toprule
\multirow{3}{*}{\tabincell{c}{Frame \\ \#Num.}}&
\multirow{2}{*}[-0.3em]{Methods} &\multicolumn{2}{c}{Vimeo90K~\cite{xue2017video}}  
&\multicolumn{2}{c}{V-BM4D~\cite{maggioni2012video}} \\
\cmidrule{3-6}
&	&PSNR & SSIM 	&PSNR & SSIM\\
	\midrule
\multirow{2}{*}{1}
%

&Blocky &31.99& 0.9179&  29.38 &  0.8302 \\


&EDSR\_DB~\cite{lim2017enhanced}& \second{32.87} & \second{0.9319}  & 29.66 &0.8362   \\

\midrule
\multirow{4}{*}{7}

&	ToFlow~\cite{xue2017video} & 32.57 & 0.9292 &29.59 & 0.8390  \\

& V-BM4D~\cite{maggioni2012video} & 32.74 & 0.9293 &\second{29.94} & \second{0.8435}\\
 	\cmidrule{2-6}
 	

&	\Ours{\_DB} & \first{33.37}   & \first{0.9388}   &\first{30.14} & \first{0.8498} \\
	
	\bottomrule
\end{tabular} 
\end{minipage}	 
\vspace{-8pt}
\end{table}

\begin{figure*}[!t]

\footnotesize
\centering
\renewcommand{\tabcolsep}{0.5pt} 
\renewcommand{\arraystretch}{0.1} 
\begin{center}
	\begin{tabular}{c ccccc}
		\multirow{2}{*}[3.05em]{	
		\includegraphics[width= 0.2\textwidth]{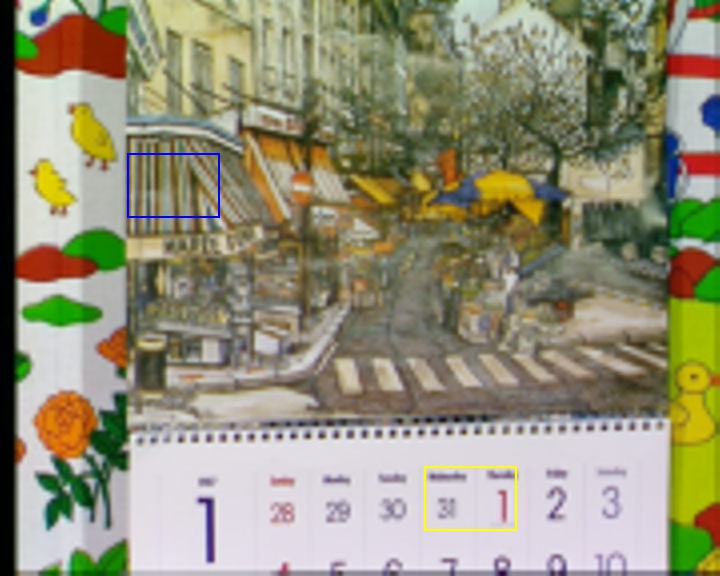}
		}

		&    
		\includegraphics[width= 0.16\textwidth]{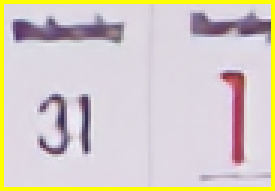}
    	&
		\includegraphics[width= 0.16\textwidth]{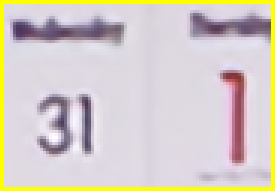}
			    	&
		\includegraphics[width= 0.16\textwidth]{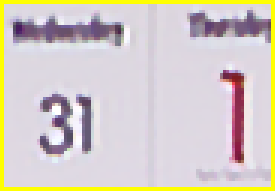}
			    	&
		\includegraphics[width= 0.16\textwidth]{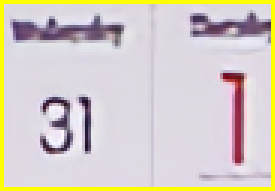}
			    	&
		\includegraphics[width= 0.16\textwidth]{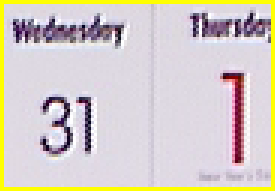}
		\\
				&    
		\includegraphics[width= 0.16\textwidth]{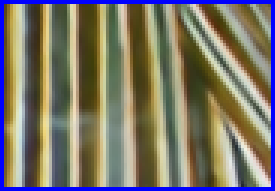}
    	&
		\includegraphics[width= 0.16\textwidth]{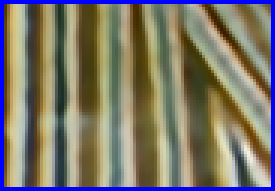}
			    	&
		\includegraphics[width= 0.16\textwidth]{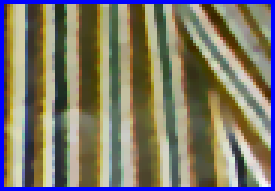}
			    	&
		\includegraphics[width= 0.16\textwidth]{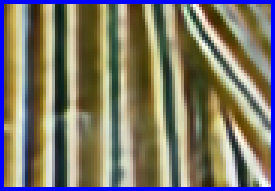}
			    	&
		\includegraphics[width= 0.16\textwidth]{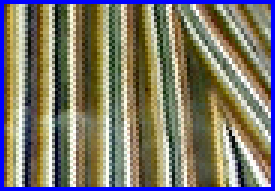}
		\\
		\addlinespace[0.1cm]
		(a) Bicubic &
		(b) EDSR~\cite{lim2017enhanced}&
		(c) ToFlow~\cite{xue2017video}&
		(d) BayesSR~\cite{liu2011bayesian} &
		(e) \Ours{\_SR} &
		(f) Ground Truth 
		\\
	\end{tabular}
\end{center}

	 \vspace{-0.4cm}
\caption{\textbf{Visual comparisons of video super-resolution methods.}}
\label{fig:super-resolution}
	 	 \vspace{-8pt}

\footnotesize
\centering
\renewcommand{\tabcolsep}{0.5pt} 
\renewcommand{\arraystretch}{0.15} 
\begin{center}
	\begin{tabular}{c ccccc}
		\multirow{2}{*}[1.31em]{	
		\includegraphics[width= 0.2\textwidth]{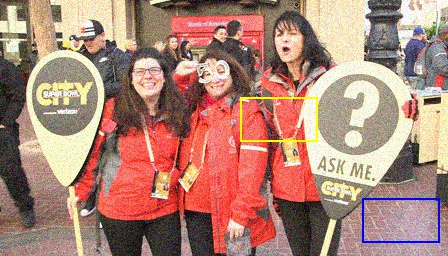}
		}
		&    
		\includegraphics[width= 0.16\textwidth]{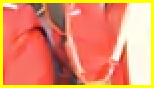}
    	&
		\includegraphics[width= 0.16\textwidth]{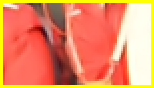}
			    	&
		\includegraphics[width= 0.16\textwidth]{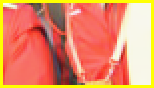}
			    	&
		\includegraphics[width= 0.16\textwidth]{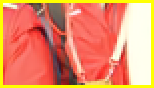}
			    	&
		\includegraphics[width= 0.16\textwidth]{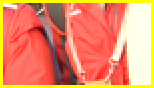}
		\\
		
			&    
		\includegraphics[width= 0.16\textwidth]{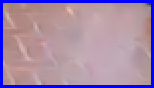}
    	&
		\includegraphics[width= 0.16\textwidth]{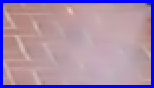}
			    	&
		\includegraphics[width= 0.16\textwidth]{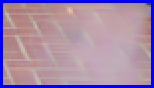}
			    	&
		\includegraphics[width= 0.16\textwidth]{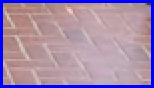}
			    	&
		\includegraphics[width= 0.16\textwidth]{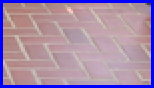}
		\\
				\addlinespace[0.1cm]

		(a) Noisy &
		(b) EDSR\_DN~\cite{lim2017enhanced}&
		(c) ToFlow~\cite{xue2017video}&
		(d) V-BM4D~\cite{maggioni2012video} &
		(e) \Ours{\_DN} &
		(f) Ground Truth 
		\\
	\end{tabular}
\end{center}

	 \vspace{-0.4cm}
\caption{\textbf{Visual comparisons of video denoising methods.}}
\label{fig:denoising}

	 	 \vspace{-8pt}

\footnotesize
\centering
\renewcommand{\tabcolsep}{0.5pt} 
\renewcommand{\arraystretch}{0.15} 
\begin{center}
	\begin{tabular}{cccccc}
		\multirow{2}{*}[0.96em]{	
		\includegraphics[width= 0.2\textwidth]{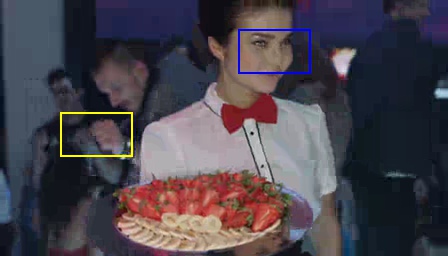}
		}
		&    
		\includegraphics[width= 0.16\textwidth]{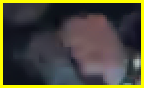}
    	&
		\includegraphics[width= 0.16\textwidth]{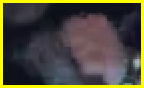}
			    	&
		\includegraphics[width= 0.16\textwidth]{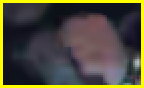}
			    	&
		\includegraphics[width= 0.16\textwidth]{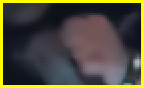}
			    	&
		\includegraphics[width= 0.16\textwidth]{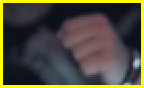}
		\\
			&    
		\includegraphics[width= 0.16\textwidth]{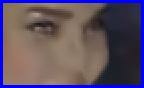}
    	&
		\includegraphics[width= 0.16\textwidth]{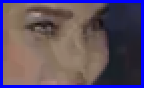}
			    	&
		\includegraphics[width= 0.16\textwidth]{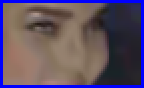}
			    	&
		\includegraphics[width= 0.16\textwidth]{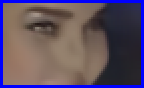}
			    	&
		\includegraphics[width= 0.16\textwidth]{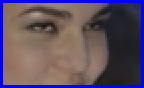}
		\\
		
		\addlinespace[0.1cm]
		(a) Blocky &
		(b) EDSR\_DB~\cite{lim2017enhanced}&
		(c) ToFlow~\cite{xue2017video}&
		(d) V-BM4D~\cite{maggioni2012video} &
		(e) \Ours{\_DB} &
		(f) Ground Truth 
		\\
	\end{tabular}
	
\end{center}

	 \vspace{-0.4cm}
\caption{\textbf{Visual comparisons of video deblocking methods.}}
\label{fig:deblocking}
	 	 \vspace{-8pt}
\end{figure*}

\subsection{Video Frame Enhancement}
We use the Vimeo90K dataset~\cite{xue2017video} to evaluate the proposed method on 
the video denoising, video super-resolution, and video deblocking tasks.
There are 7,824 sequences in the Vimeo90k test set, and each contains 7 consecutive frames.
The qualitative results for super-resolution, denoising and deblocking tasks are presented in \tabref{super-resolution}, \tabref{denoising} and \tabref{deblocking}, respectively.

\Paragraph{Super-Resolution.} 
We evaluate the proposed method on the widely used video super-resolution dataset developed by Liu~\etal~\cite{liu2011bayesian}, which is denoted by BayeSR in~\tabref{super-resolution}.
The low-resolution image distortion for both the Vimeo90K and BayesSR datasets are generated by down-sampling the original high-resolution frames at the scaling ratio of $4$ (use the MATLAB function \textit{imresize} with the \textit{bicubic} mode).
And the evaluated algorithms are to up-sample the middle frame of a sequence in Vimeo90K dataset or each frame of a video in BayesSR dataset by a factor of 4.
The DeepSR~\cite{liao2015video} and ToFlow~\cite{xue2017video} methods are CNN-based approaches for video super-resolution.
In addition, we also compare with the BayesSR~\cite{liu2011bayesian} method.
Since the single-image super-resolution (SISR) is also a well-studied task, we include 
the state-of-the-art SISR method, EDSR~\cite{lim2017enhanced}, for evaluations.

We present the quantitative results on video super-resolution in~\tabref{super-resolution}.
Our method performs favorably against the state-of-the-art approaches on both benchmark datasets.
Compared to the state-of-the-art SISR method~\cite{lim2017enhanced}, \Ours{\_SR} has fewer residual blocks and a smaller number of filters but obtains higher PSNRs on both the Vimeo90K and BayesSR datasets.
Compared to existing video super-resolution approaches~\cite{liu2011bayesian,liao2015video,xue2017video}, our method is more favorable, especially on the BayesSR dataset.
In~\figref{super-resolution}, we present the video super-resolution results.
In the first row, the EDSR~\cite{lim2017enhanced} does not 
restore the correct shape of the number ``31'' on the calendar.
The results by the ToFlow~\cite{xue2017video} and BayesSR~\cite{liu2011bayesian} methods contain artifacts and blurry pixels. 
In contrast, the proposed \Ours{\_SR} model is able to restore sharper video frames.

\Paragraph{Denoising.} 
%
%
We evaluate our method with the ToFlow~\cite{xue2017video} and 
V-BM4D~\cite{maggioni2012video} algorithms.
In addition, we train a single frame denoising model as the baseline.
The model architecture is the same as the EDSR network except that the input images are with noise instead of low-resolution, and referred to as EDSR\_DN.
We evaluate on the Vimeo90k test set as well as the dataset developed by Maggioni~\etal~\cite{maggioni2012video}.
%
%
For the denoising experiments, we add Gaussian noise with $\sigma = 20$ to synthesize noisy input frames.
%
%

The quantitative results for video denoising are presented in~\tabref{denoising}.
Our method performs well on both datasets.
The PSNR gains of \Ours{\_DN} over the second best method are 1.24dB and 1.95dB on the Vimeo90K and V-BM4D datasets, respectively.
In~\figref{denoising}, the fine textures on the clothes and street are not well restored by the EDSR\_DN, ToFlow, and V-BM4D methods.
In contrast, our \Ours{\_DN} preserves these textures well.

\Paragraph{Deblocking.}
For the video deblocking task, we use the same videos as in the denoising task.
The images encoded by the widely used H.264~\cite{wiegand2003overview}  standard may generate blockiness due to the block-based approach.
We use the FFmpeg software to encode the images in the Vimeo90K and V-BM4D datasets with \textit{libx264} and the quality parameter $qp$ of $37$, and disable the in-loop deblocking of the codec.
We compare the proposed algorithm with the EDSR\_DB~\cite{lim2017enhanced}, ToFlow~\cite{xue2017video} and V-BM4D~\cite{maggioni2012video} methods.

The quantitative evaluation results for video deblocking are presented in~\tabref{deblocking}.
Overall, the proposed model performs favorably against all the evaluated algorithms. 
%
In~\figref{deblocking}, the blocky regions around the hand and eye are sufficiently reduced by both \Ours{\_DB} and V-BM4D~\cite{maggioni2012video} methods.
The ToFlow~\cite{xue2017video} and EDSR\_DB schemes, however, do not reduce the blocky pixels well.

\section{Conclusions}
\label{sec:C} 
In this work, we propose the motion estimation and motion compensation driven neural network for learning video frame interpolation and enhancement.
Our model exploits the merits of the MEMC framework to handle large motion as well as the data-driven learning-based methods to extract effective features. 
Two network layers, namely the adaptive warping layer and flow projection layers, are proposed to tightly integrate all the sub-networks to make our model end-to-end trainable.
The generalized motion compensated alignment of the proposed MEMC framework enables it to be extended to various video enhancement tasks such as video super-resolution, denoising, and deblocking.
Quantitative and qualitative evaluations on the various benchmark datasets show that the proposed methods perform favorably against the state-of-the-art algorithms in video interpolation and enhancement.
%


\ifCLASSOPTIONcaptionsoff
  \newpage
\fi

\bibliographystyle{IEEEtran}
\bibliography{tpami18_frame_interpolation_v9} 

\begin{IEEEbiography}[{\includegraphics[width=1in,height=1.25in,clip,keepaspectratio]{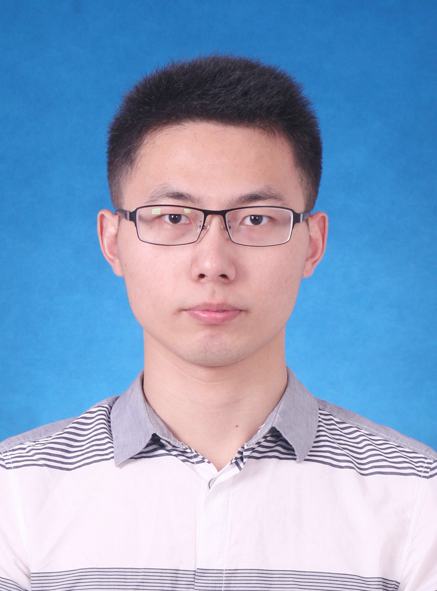}}]{Wenbo Bao}
is a Ph.D. candidate of Electrical Engineering with the Institute of Image Communication and Network Engineering, Shanghai Jiao Tong University, Shanghai, China.
He received the B.S. degree in Electronic Information Engineering from Huazhong University of Science and Technology, Hubei, China, in 2014.
His research interests include computer vision, machine learning, and video processing.
\end{IEEEbiography}

\begin{IEEEbiography}[{\includegraphics[width=1in,height=1.25in,clip,keepaspectratio]{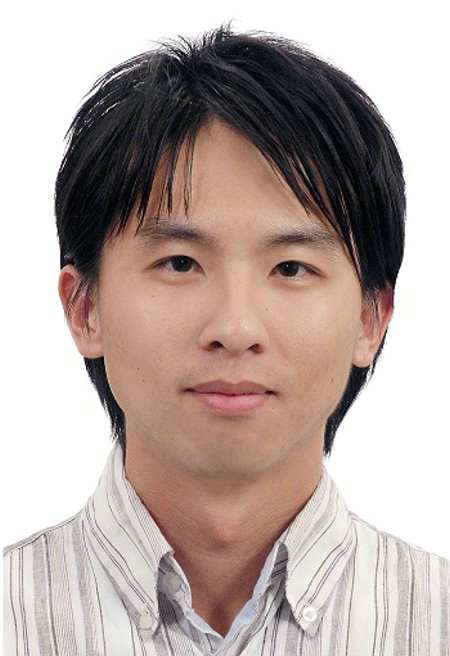}}]{Wei-Sheng Lai}
is a Ph.D. candidate of Electrical
Engineering and Computer Science at the
University of California, Merced, CA, USA. He
received the B.S. and M.S. degree in Electrical
Engineering from the National Taiwan University,
Taipei, Taiwan, in 2012 and 2014, respectively.
His research interests include computer vision,
computational photography, and deep learning.
\end{IEEEbiography}

\begin{IEEEbiography}[{\includegraphics[width=1in,height=1.25in,clip,keepaspectratio]{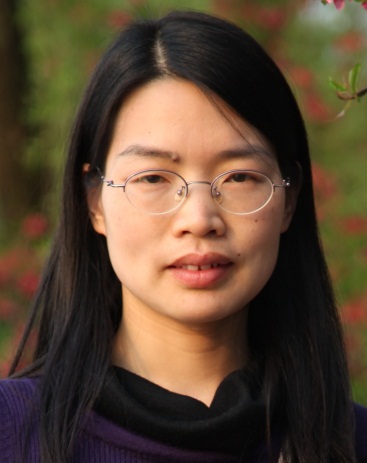}}]{Xiaoyun Zhang}
is an associate professor in Electrical Engineering at Shanghai Jiao Tong University.  
She received the B.S. and M.S. degrees in applied mathematics from Xian Jiaotong University in 1998 and 2001, respectively, and the Ph.D. degree in pattern recognition from Shanghai Jiao Tong University, China, in 2004. 
Her Ph.D. thesis has been nominated as National 100 Best Ph.D. Theses of China. 
Her research interests include computer vision and pattern recognition, image and video processing, digital TV system. 
\end{IEEEbiography}


\begin{IEEEbiography}[{\includegraphics[width=1in,height=1.25in,clip,keepaspectratio]{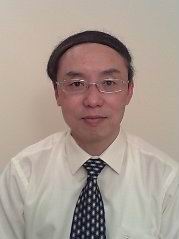}}]{Zhiyong Gao}
is a professor in Electrical Engineering with Shanghai Jiao Tong University.  
He received the B.S. and M.S. degrees in electrical engineering from the Changsha Institute of Technology, Changsha, China, in 1981 and 1984, respectively, and the Ph.D. degree from Tsinghua University, Beijing, China, in 1989. 
From 1994 to 2010, he took several senior technical positions in England, including a Principal Engineer with Snell and Wilcox, Petersfield, U.K., 
a Video Architect with 3DLabs, Egham, U.K., 
a Consultant Engineer with Sony European Semiconductor Design Center, Basingstoke, U.K.,
and a Digital Video Architect with Imagination Technologies, Kings Langley, U.K.
His research interests include video processing, video coding, digital TV, and broadcasting.
\end{IEEEbiography}

\begin{IEEEbiography}[{\includegraphics[width=1in,height=1.25in,clip,keepaspectratio]{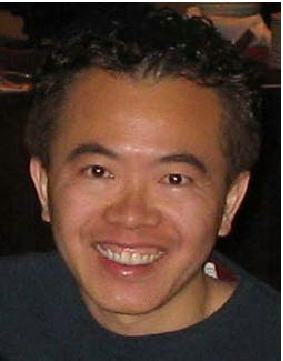}}]{Ming-Hsuan Yang}
is a professor in Electrical
Engineering and Computer Science at University
of California, Merced. He received the Ph.D.
degree in computer science from the University
of Illinois at Urbana-Champaign in 2000.
Yang serves as a program co-chair of IEEE International Conference on Computer Vision (ICCV) in 2019, program co-chair of Asian Conference on Computer Vision (ACCV) in 2014, and general co-chair of ACCV 2016.
Yang served as an associate editor of the IEEE
Transactions on Pattern Analysis and Machine
Intelligence from 2007 to 2011, and is an associate
editor of the International Journal of Computer
Vision, Image and Vision Computing and
Journal of Artificial Intelligence Research. He
received the NSF CAREER award in 2012, the Senate Award for Distinguished
Early Career Research at UC Merced in 2011, and the Google
Faculty Award in 2009. 
He is a Fellow of the IEEE and a Senior Member of the ACM.
\end{IEEEbiography}

\end{document}